\pdfoutput=1

\documentclass[11pt]{article}

\usepackage[final]{acl}

\usepackage{times}
\usepackage{latexsym}

\usepackage[T1]{fontenc}

\usepackage[utf8]{inputenc}

\usepackage{thmtools,thm-restate}
\usepackage{colortbl}
\usepackage{amssymb}
\usepackage{hyperref}       
\usepackage{graphicx}
\usepackage{threeparttable}
\usepackage[utf8]{inputenc} 
\usepackage[T1]{fontenc}    
\usepackage{url}            
\usepackage{nicefrac}       
\usepackage{microtype}      
\usepackage{xcolor}         
\usepackage{subcaption}
\usepackage{bbm}
\usepackage{authblk}
\usepackage{enumitem}

\usepackage{algorithm}
\usepackage{algpseudocode} 
\usepackage{amsmath} 
\usepackage{amsfonts} 
\usepackage{array} 
\usepackage{bbding}
\usepackage{booktabs} 
\usepackage{caption} 
\usepackage{cleveref}
\usepackage{fancyvrb}
\usepackage{float}
\usepackage{listings}
\usepackage{makecell}
\usepackage{mathtools} 
\usepackage{multicol}
\usepackage{multirow}
\usepackage{pifont}
\usepackage{tabularx}
\usepackage{wrapfig}
\usepackage{xspace}
\usepackage{fix-cm}

\usepackage{forloop}

\NewDocumentCommand{\var}{O{s} m O{}}{%
  \ensuremath{#1_{#2}^{#3}}
}
\usepackage{siunitx}

\newcommand{\commentout}[1]{}

\definecolor{light-gray}{gray}{0.80}

\newcommand\fref{Fig.~\ref}

\newcommand\sref{\S~\ref}

\usepackage{amsthm}

\usepackage{amsthm}

\newtheorem{mytheorem}{Theorem}
\newtheorem{assumption}[mytheorem]{Assumption}
\newtheorem{lemma}[mytheorem]{Lemma}
\newtheorem{remk}[mytheorem]{Remark}

\newcommand{\ho}{H$_2$O\xspace}
\newcommand{\snap}{SnapKV\xspace}
\newcommand{\name}{MiniKV\xspace}

\definecolor{darkred}{rgb}{0.7, 0, 0}
\definecolor{darkblue}{rgb}{0, 0, 0.9}
\usepackage[skip=2pt]{caption}
\usepackage{setspace}
\usepackage{enumitem}
\usepackage{titlesec}

\setlist{nosep} 
\newcommand{\setvspace}[2]{%
  #1 = #2
  \advance #1 by -1\parskip}

\makeatletter 
\def\thm@space@setup{%
  \thm@preskip=3pt
  \thm@postskip=\thm@preskip 
}
\makeatother

\setlist[itemize]{noitemsep, topsep=0pt}

\makeatletter
\g@addto@macro\normalsize{%
  \setlength\abovedisplayskip{1pt}
  \setlength\belowdisplayskip{1pt}
  \setlength\abovedisplayshortskip{1pt}
  \setlength\belowdisplayshortskip{1pt}
}
\makeatother

\makeatletter 
\renewenvironment{proof}[1][\proofname]{\par
  \vspace{1pt}
  \pushQED{\qed}%
  \normalfont
  \topsep0pt \partopsep0pt 
  \trivlist
  \item[\hskip\labelsep
        \itshape
    #1\@addpunct{.}]\ignorespaces
}{%
  \popQED\endtrivlist\@endpefalse
  \addvspace{3pt plus 3pt} 
}
\makeatother 

\title{\name: Pushing the Limits of 2-Bit KV Cache via Compression and System Co-Design for Efficient Long Context Inference}

\author{
 \textbf{Akshat Sharma},
 \textbf{Hangliang Ding}\thanks{Work done while intern at UIUC. Correspondence to: Minjia Zhang (minjiaz@illinois.edu) \\
 Project Homepage: \href{https://supercomputing-system-ai-lab.github.io/projects/minikv/}{https://supercomputing-system-ai-lab.github.io/projects/minikv/}
 },
 \textbf{Jianping Li},
 \textbf{Neel Dani},
 \textbf{Minjia Zhang}
\\
 SSAIL Lab, University of Illinois at Urbana-Champaign
\\
\texttt{\{akshat7, jli199, neeld2, minjiaz\}@illinois.edu}
\\
\texttt{pianoqwz@gmail.com}
}

\begin{document}
\maketitle
\begin{abstract}
State-of-the-art 2-bit KV cache quantization techniques achieve excellent results in accelerating LLM inference while retaining accuracy on long context tasks. However, further pushing the compression ratio fails to deliver performance gains.  
In this work, we revisit these approaches by considering, additionally, adaptive KV methods that retain LLM accuracy with only a subset of KV states. This leads us to propose a method based on 2-bit KV cache quantization with adaptive KV policies. In addition, we take an algorithm and system co-design approach by developing hardware-friendly kernels to accelerate LLM inference while making \name compatible with existing memory-efficient attention techniques such as FlashAttention, effectively translating algorithmic improvements into system performance gains. Experiments on a wide range of long context tasks show that \name effectively achieves $>$80\% KV cache compression while retaining accuracy, outperforming state-of-the-art methods while achieving excellent latency, throughput, and memory consumption improvements in long context inference.
\end{abstract}

\section{Introduction}
\label{sec:intro}

Large language models (LLMs) have exhibited unique capabilities, such as instruction following, reasoning, and inference time scaling~\cite{gpt-o1,deepseek2025deepseek}. However, efficiently serving LLMs is still a pressing concern. One of the main LLM inference bottlenecks is the consumption of KV cache memory, which consumes memory in addition to widely studied bottlenecks such as model sizes~\cite{gptq,awq}.

To address this challenge, one of the prevailing practices is to quantize the KV cache~\cite{vllm-quantized-kv,tensorrt-quantized-kv}. Studies show that FP8/INT8 or even 4-bit quantization can be achieved for KV cache compression while preserving accuracy~\cite{flexgen,llm-qat,no-token-left}. State-of-the-art approaches, such as KIVI and KVQuant~\cite{kivi, kvquant}, show that the KV cache can be effectively quantized to sub 4-bit, e.g., 2 bits, while preserving most accuracy. However, further pushing down the compression ratio (e.g.,$<$2 bits) leads to a significant accuracy loss. 

In a separate line of research in the community, numerous work have explored \emph{adaptive KV}, where the LLM selects a small subset of KV states based on their importance~\cite{h2o,attention-sink,fastgen,scissorhands}. Recent advances also introduce head-specific adaptive KV~\cite{fastgen,duoattention,retrieval-head} and layer-specific adaptive KV~\cite{pyramidkv,DMC,d2o} with the goal of evicting or merging KV pairs without compromising overall performance. However, following the work of \cite{h2o}, few studies have included studies on how adaptive KV policies work on quantized KV cache, despite quantized KV is widely used in practice~\cite{kivi-hf}. Moreover, for long context inference, where KV cache memory becomes the major bottleneck, few adaptive KV work manage to achieve a compression ratio that exceeds 50\% while maintaining accuracy in long context tasks~\cite{li2024snapkv,tang2024quest}. 

\begin{figure*}[!ht]
    \centering
    \includegraphics[width=\linewidth]{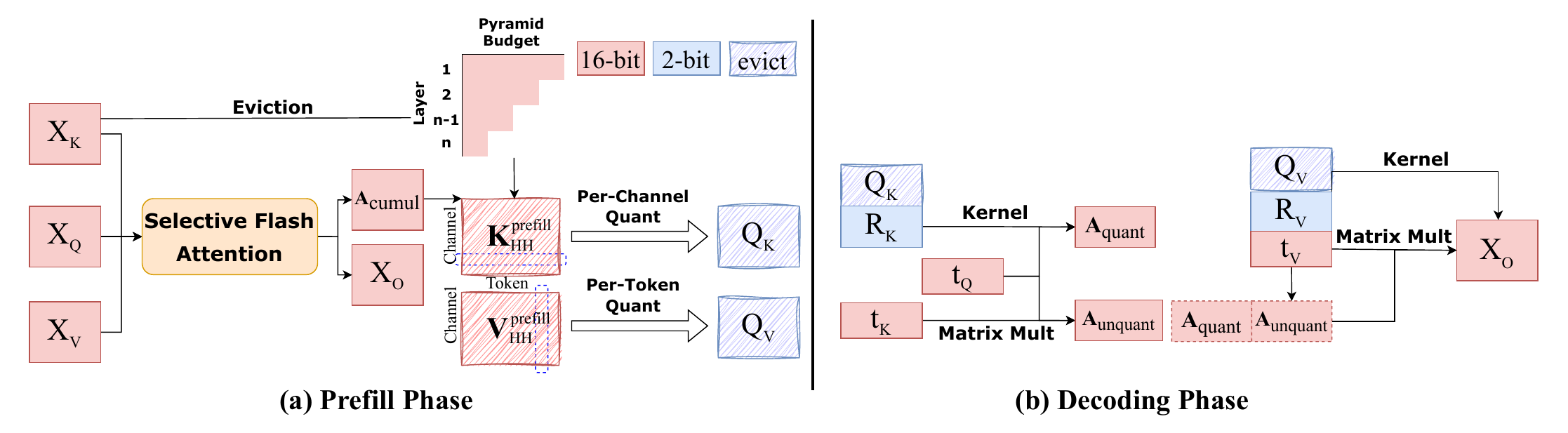}
    \caption{An overview of \name. 
    Tensors colored \textcolor{red}{red}/\textcolor{blue}{blue} indicate \textcolor{red}{16-bit}/\textcolor{blue}{2-bit} representation, and shaded tokens are evicted during inference.
    During the prefill phase, we employ pyramid KV with rectified token selection policy across layers to identify a sparse set of important tokens. For all the important tokens, we employ sub-channel Key quantization and per-token Value quantization to minimize the quantization errors while maintaining a compact KV cache data layout without introducing any irregular operations. To address the incompatibility issue between score-based KV pair selection policies and memory-efficient system optimizations such as FlashAttention, we develop a two-pass Triton-based \emph{selective flash-attention kernel} to output both the representation $X_O$ and the cumulative attention map $A_{\text{cumul}}$, while still keeping the memory consumption of the attention calculation linear with respect to the sequence length. 
    During decoding, we use a \emph{fused unpacking and multiplication} kernel to compute both the attention map between the new Query token $t_Q$ and the quantized Keys, as well as the product between the attention map and the quantized Values.
    }
    \label{fig:algo_figure}
\end{figure*}

These two points of view (quantized KV and adaptive KV) consider the extreme sides of the spectrum of optimization points for KV cache memory. However, there has been very little work exploring how to consolidate these two lines of work to maximize the KV cache memory savings. The conventional wisdom is that these techniques can be \emph{combined}. However, existing work aiming to combine 4-bit quantization and adaptive KV shows that combining these techniques leads to non-trivial interactions~\cite{q-hitter}, which need to be reasoned through carefully for good performance. In this paper, we address the following question:
\emph{How should 2-bit KV cache quantization techniques be combined with adaptive KV policies to maximize the inference speed of LLMs given a memory budget while retaining high model accuracy in long context inference?}

To answer the question, we revisit existing approaches on ultra low-bit quantized KV (e.g., 2-bit) and adaptive KV, together with a compression system co-design perspective, which is unexplored so far. Our findings led us to develop \textbf{\name}, which effectively compresses the KV cache through a synergistic combination of 2-bit quantization and adaptive KV to achieve minimal accuracy loss in long-context tasks while maximizing the compression ratio. Specifically, on the algorithm side, we employ subchannel-wise key and token-wise value quantization, as well as pyramid KV with rectified token selection policy across layers to significantly push the KV compression ratio while keeping the algorithm still hardware-friendly without introducing any irregular computation. On the system side, we develop a two-pass Triton~\cite{triton} kernel together with native fused kernels to accelerate the inference latency while resolving the incompatibility limitation from the attention score-based eviction policy and memory-efficient attention system optimizations such as FlashAttention~\cite{flash-attention}. Consequently, the resulting system maximizes the compression ratio on the KV cache while obtaining high accuracy and hardware efficiency in long context inference. 
 
To validate the approach, we compare \name with existing KV cache compression techniques such as H2O, SnapKV, and Q-Hitter, across three major models in LongBench datasets. The results show that \name effectively achieves 86\% KV cache compression while retaining comparable accuracy on LongBench, outperforming state-of-the-art methods. Furthermore, \name enables prompt lengths up to 44K tokens and a maximum throughput that is 48\% higher than its strongest baseline on a single NVIDIA A100 GPU. To our knowledge, our work is the first to show that it is possible to achieve significantly $>$50\% KV cache reduction through compression and system co-design while achieving high batch size $\geq 1$ throughput on long context tasks.

\section{Related Work}
\label{sec:background}

Numerous efforts have been made to improve the KV cache efficiency of LLMs. Among them, quantization has been a prevailing technique employed in deployment to overcome KV memory overhead without retraining~\cite{vllm-quantized-kv,tensorrt-quantized-kv}. 
Many research has shown that FP8/INT8/INT4 quantization can be achieved for KV cache while preserving accuracy~\cite{kvquant,flexgen,llm-qat,no-token-left,q-hitter}. However, further pushing the quantization limit to under 4-bit, e.g., 2-bit, leads to major performance loss. More recently, researchers have proposed advanced quantization techniques, such as KIVI~\cite{kivi}, to quantize KV cache into 2-bit without major loss in accuracy. While being effective, it still has one major limitation: its effectiveness against adaptive KV policies and its implication on system performance has not yet been studied. Our results indicate that it is nontrivial to use 2-bit quantized KV together with adaptive KV policies in conjunction while achieving high compression ratio, accuracy, and system efficiency in long context inference, simultaneously.   


Adaptive KV policies have also gained interest within the community, leading to various algorithms~\cite{h2o,attention-sink,scissorhands,fastgen,d2o,retrieval-head, pyramidkv,Pyramid,li2024snapkv,minicache,CLA,tang2024quest}. 
However, many of those works either do not focus on long context inference~\cite{h2o,attention-sink,scissorhands,fastgen}, where the KV cache pressure is the most prominent, or introduce irregular operations or auxiliary scores that are not hardware-friendly (e.g., batch size $>$1 with FlashAttention enabled)~\cite{fastgen,retrieval-head,d2o,tang2024quest}. Finally, most adaptive KV methods struggle to exceed a 50\% compression rate in long context inference~\cite{h2o,li2024snapkv,tang2024quest}, suggesting that solely identifying important tokens may have limited improvements for adaptive KV. 
Complementary to this line of work, our goal is to improve the compression ratio of KV cache via revising ultra-low quantized KV (e.g., 2-bit) with adaptive KV policies, with an eye toward system co-design to maximize the performance of LLM inference. We empirically show that this path can be more memory-efficient, especially on long context tasks. We provide a detailed summary of the comparison between \name and previous approaches in Appendix~\ref{appendix:copmarison}. 



\section{Method}
\label{sec:design}

In this section, we first focus on the compressibility of ultra low-bit quantized KV cache by considering adaptive KV policies, with an eye toward being able to still keep the final solution hardware friendly, which leads to the proposed algorithm in \name. In addition, we introduce kernel optimization that addresses the composibility issue between score-based adaptive KV and memory-efficient attention implementation such as FlashAtttention. 

\subsection{Revisiting 2-Bit Quantized KV with Adaptive KV Policies}

\subsubsection{Sub-channel Key Quantization with Persistent Context Selection}
Existing KV cache quantization methods often perform per-token quantization (i.e., the scaling factor and zero point are shared by elements in the same token)~\cite{flexgen,smoothquant}.
However, it has been observed
that outliers emerge within the channel dimension of key cache~\cite{kivi,kvquant}, requiring channel-wise quantization. 


Recent works ~\cite{kvquant, kivi} observe that the data distribution within each channel shifts over generation steps, leading to inaccurate quantization. We measure and confirm the accuracy impact of inaccurate quantization on LongBench in Appendix \ref{appendix:token_wise_quant}.
To mitigate quantization error, prior work suggests fine-grained per-channel key quantization, which quantizes keys at the granularity of a small sub-channel group (e.g. 16/32 numbers).
Combining these techniques with a full KV cache is straightforward because the elements within each sub-channel group remain the same during the entire LLM generation process.

However, with adaptive KV, the elements within a sub-channel group may change after each decoding step if some tokens in the group are evicted to make space for newly generated tokens. \name solves this problem by enabling sub-channel key quantization via \emph{persistent context selection}. 
Our design for this optimization is based on the following key observation: \emph{Given a sufficiently large cache budget, the important tokens can be identified before generation and maintained persistently throughout the process.}


We found some recent inference optimization works that argue against persistent context selection \cite{tang2024quest}. However, all of these texts show that heavy hitters do not persist when using a tiny cache budget (<5\%).
We empirically verify persistent context selection by measuring the fraction of heavy hitters that persist during the entire generation phase of \ho when using a large cache budget. We observe that nearly 60-80\% of the heavy hitters selected during the prefill stage persist throughout generation. Please see Appendix \ref{appendix:persistent} for details.


Based on this observation, we choose a set of \emph{persistent heavy hitters} at the end of the prefill to quantize and not update throughout the generation phase. This allows \name to avoid re-encoding a group while keeping a low quantization error with 2-bit sub-channel quantization.

\subsubsection{Selectivity in Long Contexts: Heavy Hitters vs. Recent Window}
Prior studies observe that the accumulated attention scores of all tokens within an attention block follow a power-law distribution and claim that maintaining a tiny subset of important tokens (e.g., as low as 5\%) with the highest accumulated attention score is sufficient to maintain precision~\cite{h2o}. However, this observation has not been carefully examined in long contexts. 

\begin{figure}[!ht]
    \includegraphics[width=1.00\linewidth]{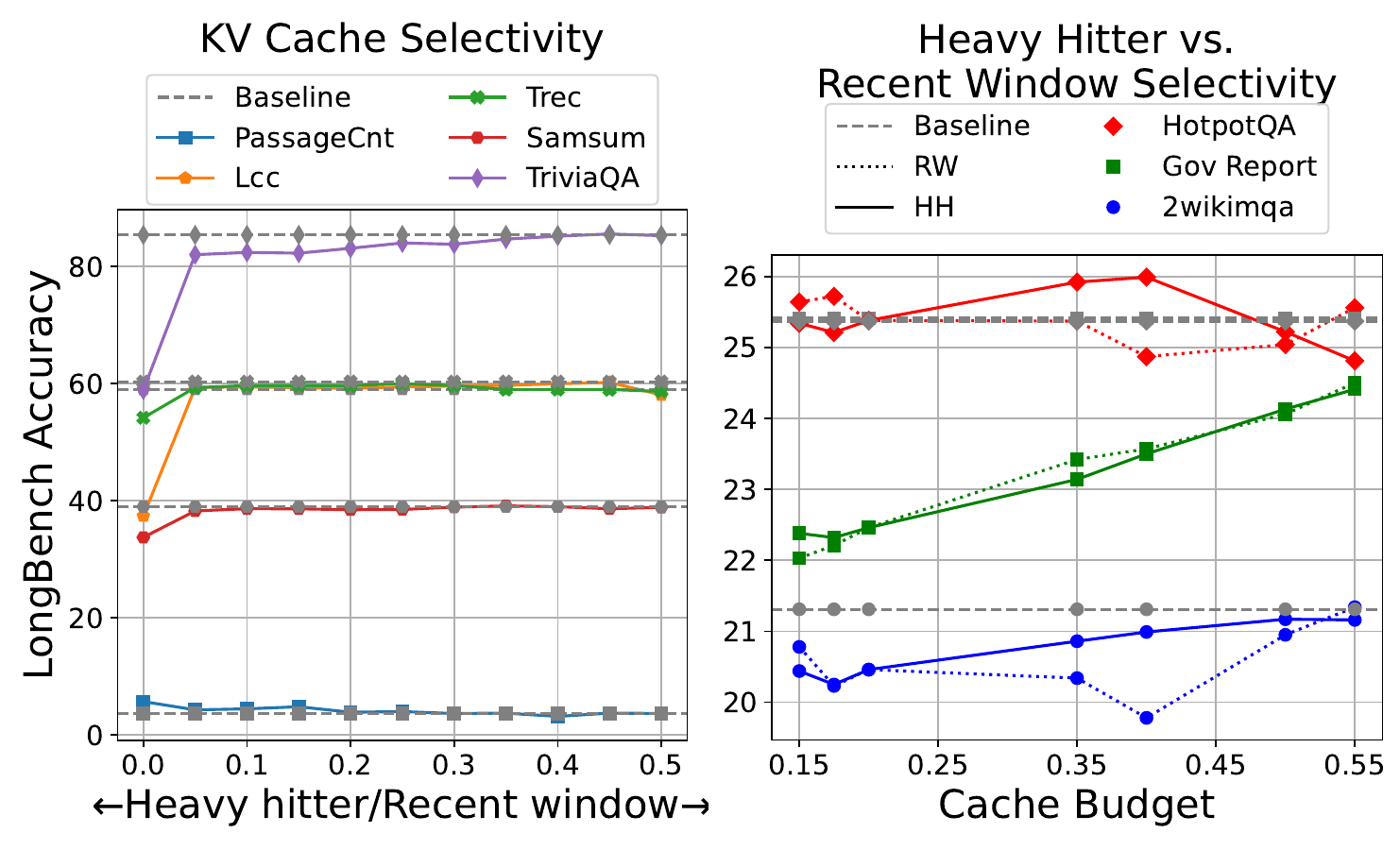}
    \caption{(Left) \ho with different recent window/heavy hitter budget: We fix the total cache budget to $50\%$ and vary the heavy hitter and recent window budget. 
    (Right) \ho with different recent window/heavy hitter budget: The heavy hitter/recent window cache budget is fixed at $10\%$ and the recent window/heavy hitter budget is increased from $5\%$ to $45\%$. The dotted/solid lines indicate variable recent window/heavy hitter budget. 
    }
    \label{fig:merged_kv_cache_selectivity}
\end{figure}

We observe that using a highly limited memory budget (e.g., 20\%), existing solutions such as \ho~\cite{h2o} and SnapKV~\cite{li2024snapkv} have a significant performance drop in long context tasks, which motivates us to revisit the selectivity of adaptive KV methods. 
First, we assess if the model retains performance using only the recent window (RW) or heavy hitters (HH), we examine the KV cache's selectivity towards RW/HH.
The cache budget is described as the percent of the prompt tokens retained, i.e. an RW/HH budget of $(\alpha_{RW}, \alpha_{HH})$ and an input prompt of length $l_{\text{prompt}}$ tokens indicate that $(\alpha_{RW} \cdot l_{\text{prompt}}, \alpha_{HH} \cdot l_{\text{prompt}})$ tokens are maintained as the RW and HH respectively.
We fix the total cache budget to $50\%$ and distribute it among the RW and HH, i.e., RW/HH budget of $(\alpha_{RW}, \alpha_{HH}) = (0\%, 50\%), (10\%, 40\%), (20\%, 30\%)$, and so on. \fref{fig:merged_kv_cache_selectivity} (left) reveals an interesting aspect of the KV cache selectivity: The model performs better on some datasets with more HH (on Passage Count) and on some with a longer RW (on TriviaQA). More importantly, using solely RW or HH leads to a catastrophic accuracy drop in certain tasks (in Lcc and TriviaQA). 
This indicates that to have a robustly optimized KV cache selection policy, the model needs to maintain at least a critical percentage of HH/RW (e.g., 5-10\%) to avoid a significant accuracy drop.

Next, we investigate the selectivity between RW and HH by varying the KV cache budget. In particular, we fix the RW size (e.g., $10\%$ of the prompt length) while varying the HH set size, and vice versa. Interestingly, as seen in \fref{fig:merged_kv_cache_selectivity} (right), we observe that there appears to be \emph{no common trend} across datasets as to whether increasing the size of the RW vs. the HH set significantly improves the selectivity of KV states on long context tasks. In fact, either HH or RW allow adaptive KV to achieve accuracy comparable to the full KV cache baseline. Furthermore, unlike previous findings, which suggest that high levels of eviction (80-95\%) do not decrease model accuracy~\cite{h2o}, we find that as the sequence length increases, maintaining accuracy under the same KV cache size budget becomes challenging (please see Appendix \ref{appendix:h2o_snap_longbench} for more details). However, low and medium levels of eviction (e.g., 50\%) are still possible.

\textbf{Insight.} Our experiments suggest that high levels of KV cache eviction significantly degrade LLM's performance on long context tasks. However, medium levels of eviction can still retain comparable model accuracy. Even at medium levels, the model needs to maintain a critical percentage of both heavy hitters and recent window tokens.

\subsubsection{Layer-Specific Selectivity: Uniform, Variance, or Pyramid?}
\label{sec:layer_wise_disc}
Inspired by recent works on layer-wise KV cache compression ~\cite{pyramidkv,minicache,d2o}, we investigate several layer-specific KV cache selection strategies that allocate variable KV cache budgets across model layers.



\begin{itemize}
    \item \textbf{Uniform allocation}: This policy has been used in multiple previous studies~\cite{h2o,attention-sink,scissorhands}, where all layers have the same KV cache budget.
    \item \textbf{Variance-based allocation}: \
    Similar to \cite{d2o}, we use the variance of the cumulative attention map to determine the layer-wise KV cache budget. Lower layers exhibit smaller variances, making token eviction difficult. 
    We examine two policy variations: \emph{Var-prop}, allocating KV cache per layer proportionally to variance, and \emph{Var-inv}, allocating it inversely proportional to variance.
    \item \textbf{Pyramid-like allocation}: 
    \label{layerwise:pyramid}
    Introduced in \cite{pyramidkv}, this strategy adjusts the heavy hitter cache budget across layers by allocating more cache in lower layers and less in higher ones. The cache budget for the intermediate layers is determined through linear interpolation.
\end{itemize}



\begin{figure}[ht!]
    \centering
    \includegraphics[width=.8\linewidth]{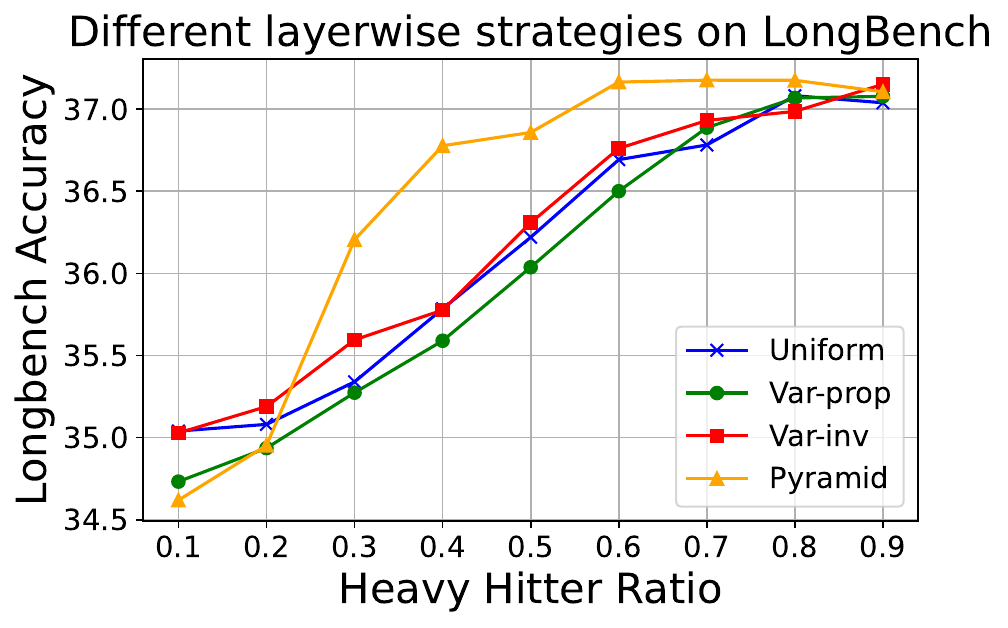}
    \caption{Performance of layer-wise KV cache allocation policies. The \textit{Pyramid} policy works best, particularly at medium levels of eviction.}
    \label{fig::layer_curve}
\end{figure}

\noindent
In our experiments, we observe that the \emph{Pyramid} policy achieves much better accuracy than the other policies, especially with medium levels of eviction, shown in \fref{fig::layer_curve}.

\subsection{Memory-Efficient Fused Selective Attention Kernels}
\label{subsec: kernel}

Despite ongoing advancements, many adaptive KV studies predominantly use attention scores as a criterion to determine which tokens should be evicted ~\cite{h2o,scissorhands,snap,pyramidkv}. 

While showing promising results in reducing the KV cache size, these \emph{attention-score-driven} methods are not aligned with memory-efficient transformer system optimizations. In particular, these methods rely on accessing the attention matrix $A$, whose size grows quadratically with the sequence length. FlashAttention~\cite{flash-attention} performs the attention process without materializing the attention matrix. Therefore, to the best of our knowledge, no prior adaptive KV works with FlashAttention enabled, hindering their memory savings on long sequences.


To address the challenge, this part introduces our memory-efficient Triton kernel implementation for \name's prefill phase, which simultaneously returns the following two outputs with linear memory complexity: 
(1) a weighted sum of the value tensors $\textcolor{darkred}{X_O}$, same as FlashAttention, and (2) cumulative attention score $\textcolor{darkred}{A_{cumul}}$
along each column~\footnote{Variables marked in \textcolor{darkred}{Red}/\textcolor{darkblue}{Blue} indicate tensors in \textcolor{darkred}{FP16}/\textcolor{darkblue}{INT2} precision.}. 
Despite being a simple task when memory is not a constraint, implementing such a kernel with linear memory complexity is challenging. The difficulty arises because $\textcolor{darkred}{A_{cumul}}$ requires summing the attention values for each token position, i.e., along the columns of the attention matrix.
FlashAttention reduces memory usage by employing row-wise tiling, which avoids storing large intermediate attention matrices. However, this row-wise tiling means that different rows are processed in parallel, leading to a race condition when summing the attention scores column-wise. To prevent this race condition, atomic add instructions are needed, which significantly slow down the kernel execution speed. 

We solve this by introducing a two-pass kernel implementation. 
In the first-pass of the kernel, we follow FlashAttention to compute the weighted sum of the value tensors and save the intermediate LSE (Log Sum Exponential) value. To efficiently operate on data in shared memory, we tile the input query tensor into row blocks of size KBlockM. 
Within each row block, the key tensor is subdivided into tile blocks of size KBlockN. 
Each row and column block calculates the tiled attention map \( P^{KBlockM \times KBlockN} \). With this product of the query and key tensors and the corresponding tile from the value tensor, we follow FlashAttention's online softmax reduction to compute the weighted V block write it back.
We aggregate the LSE value per row into an additional buffer of size $[\text{batchSize}, \text{headDim}, \text{seqLen}]$.

\begin{figure}[t]
    \centering
    \includegraphics[width=1.0\linewidth]{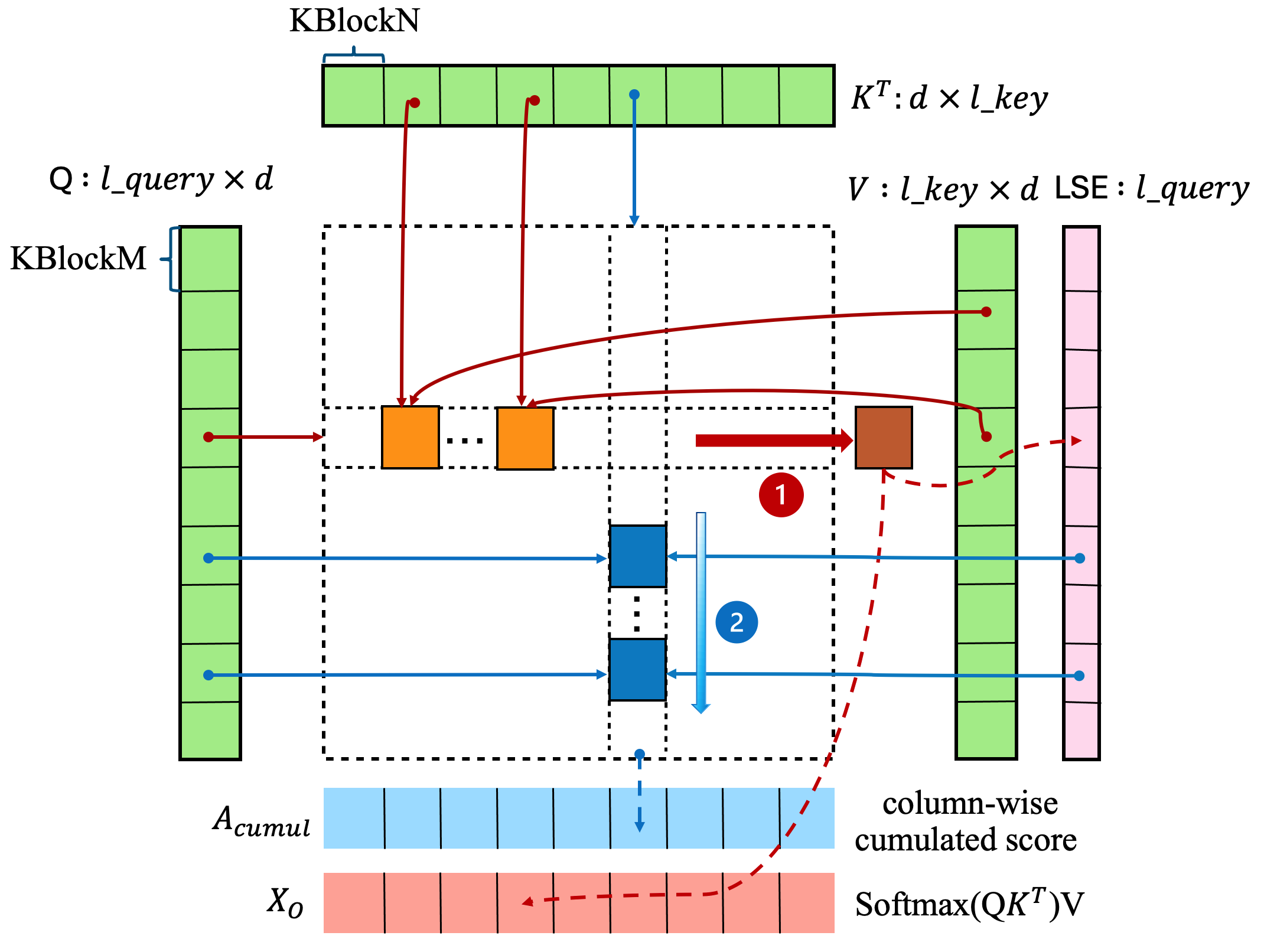}
    \caption{Two-pass kernel parallelism: In the first pass, we choose different row blocks running in parallel to compute the weighted sum of value tensors. At the same time, each row updates its max and sum and saves it to LSE. Then it switches to processing column blocks in parallel during the second pass. For each column, it recomputes $QK^T$ and normalizes it with the corresponding LSE value. From top to bottom, each column accumulates the sum and writes the result to $\textcolor{darkred}{A_{\text{cumul}}}$.}
    \label{fig:kernel-second-pass}   
\end{figure}

For the second-pass, we run different columns in parallel to compute a sequential sum of attention weights per token.
As shown in \fref{fig:kernel-second-pass}, we iteratively recompute the  $QK^T$ value and use the LSE values to normalize it. From top to bottom, we accumulate the sum column-wise and save it to the corresponding position in \(\textcolor{darkred}{A_{\text{cumul}}}\).

In summary, any memory buffers that we allocate over FlashAttention scale linearly with sequence length, i.e., LSE requires \( O(l_{\text{query}}) \) and \(\textcolor{darkred}{A_{\text{cumul}}}\) requires \( O(l_{\text{key}}) \) memory. 




\subsection{\name Algorithm} 

Based on the aforementioned observations and optimizations, we employ compression and system co-design for \name, as shown in Algorithm~\ref{algo:code}. In the prefill stage, \name uses the fused selective flash-attention kernel (\sref{subsec: kernel}) to obtain aggregated attention scores $\textcolor{darkred}{A_{\text{cumul}}}$. Based on the attention score, \name selects the subset of KV states that has the highest attention score at the end of the prefill stage (denoted as $\textcolor{darkred}{K_{HH}^{\text{prefill}}}, \textcolor{darkred}{V_{HH}^{\text{prefill}}}$).
The tokens retained are compressed to INT2 representations.
We use a separate high-performance compression kernel provided by \cite{kivi} to apply bit shift to pack 16 INT2 scalar values from selected KV states into an INT32 tensor. 
The key/value tokens are quantized along the channel/token dimension.
The results at the end of the prefill phase are the quantized key/value representation ($\textcolor{darkblue}{Q_K, Q_V}$, stored in packed INT32 tensors) and the quantization zero-point and scale (stored in FP16 tensors).

During each decoding step, \name dequantizes the quantized KV cache ($\textbf{q}^{-1}(\textcolor{darkblue}{Q_K, Q_V})$) and uses the dequantized key states along with the new key and query token $(\textcolor{darkred}{t_K, t_Q})$ for attention calculation. 
Once the attention map $(\textcolor{darkred}{A})$ is obtained, the dequantized values states ($\textbf{q}^{-1}(\textcolor{darkblue}{Q_V})$) and the new value token $(\textcolor{darkred}{t_V})$ are multiplied by $(\textcolor{darkred}{A})$ to compute the output of the attention layer $(\textcolor{darkred}{t_O})$.
\name fuses the dequantization operations with subsequent matrix multiplications to reduce kernel launch overhead and global memory accesses, leading to latency reduction.

Inspired by KIVI~\cite{kivi}, we use a streaming buffer for both key and value states during the decoding stage, so that newly generated key/value caches are first stored in FP16 (indicated by ($\textcolor{darkred}{R_K, R_V}$)).
These tokens are compressed every $n_r$ step. This saves repeated kernel launch overhead for quantization while maintaining at most $n_r$ KV tokens in FP16 during generation. 

\begin{algorithm}[t]
\caption{The \name Algorithm, \textcolor{darkred}{FP16}/\textcolor{darkblue}{INT2}}
\label{algo:code}
\small
\begin{algorithmic}[1]

\Require Input $\textcolor{darkred}{X_P} \in \mathbb{R}^{l_{\text{prompt}} \times d}$
\State $\textcolor{darkred}{X_Q}, \textcolor{darkred}{X_K}, \textcolor{darkred}{X_V} \gets \textcolor{darkred}{X_P} W_Q, \textcolor{darkred}{X_P} W_K, \textcolor{darkred}{X_P} W_V$
\State $\textcolor{darkred}{X_O}, \textcolor{darkred}{A_{\text{cumul}}} = \text{Selective\_flash\_attn}(\textcolor{darkred}{X_Q}, \textcolor{darkred}{X_K}, \textcolor{darkred}{X_V})$
\State $\textcolor{darkred}{K_{HH}^{\text{prefill}}}, \textcolor{darkred}{V_{HH}^{\text{prefill}}}, \#_{HH} \gets \text{Heavy\_hitters}(\textcolor{darkred}{A_{\text{cumul}}})$
\State $\textcolor{darkblue}{Q_K}, \textcolor{darkblue}{Q_V} \gets \textrm{Quant}(\textcolor{darkred}{K_{HH}^{\text{prefill}}}), \textrm{Quant}(\textcolor{darkred}{V_{HH}^{\text{prefill}}})$
\State $\text{KV Cache} \gets \textcolor{darkblue}{Q_K}, \textcolor{darkblue}{Q_V}$

\Procedure{\textbf{Decoding}}{$\textrm{KV cache, token } \textcolor{darkred}{t} \in \mathbb{R}^{1 \times d}$}
    \State $\textcolor{darkred}{t_Q}, \textcolor{darkred}{t_K}, \textcolor{darkred}{t_V} \gets \textcolor{darkred}{t} W_Q, \textcolor{darkred}{t} W_K, \textcolor{darkred}{t} W_V$
    \State $\textcolor{darkblue}{Q_K}, \textcolor{darkblue}{Q_V}, \textcolor{darkred}{R_K}, \textcolor{darkred}{R_V} \gets \text{KV cache}$
    \State $\textcolor{darkred}{R_K}, \textcolor{darkred}{R_V} \gets \text{Concat}([\textcolor{darkred}{R_K}, \textcolor{darkred}{t_K}]), \text{Concat}([\textcolor{darkred}{R_V}, \textcolor{darkred}{t_V}])$
    
    \If{$\text{len}(\textcolor{darkred}{R_K}) = n_r$}
        \State $\textcolor{darkblue}{Q_K'}, \textcolor{darkblue}{Q_V'} \gets \text{Quant}(\textcolor{darkred}{R_K}), \text{Quant}(\textcolor{darkred}{R_V})$
        \State $\textcolor{darkblue}{Q_K} \gets \text{Concat}([\textcolor{darkblue}{Q_K}, \textcolor{darkblue}{Q_K'}], \text{dim = channel})$
        \State $\textcolor{darkblue}{Q_V} \gets \text{Concat}([\textcolor{darkblue}{Q_V}, \textcolor{darkblue}{Q_V'}], \text{dim = token})$
        \State $\textcolor{darkred}{R_K}, \textcolor{darkred}{R_V} \gets \text{None}$
    \EndIf
    
    \State $\textcolor{darkred}{A} \gets \textrm{Softmax}(\text{Concat}([\mathbf{q^{-1}(}\textcolor{darkblue}{Q_K}\mathbf{)} \textcolor{darkred}{t_Q^T}, \textcolor{darkred}{R_K} \textcolor{darkred}{t_Q^T}]))$
    \State $\textcolor{darkred}{A_{\textrm{quant}}}, \textcolor{darkred}{A_{\text{unquant}}} \gets \textcolor{darkred}{A}[:-\text{len}(\textcolor{darkred}{R_K})], \textcolor{darkred}{A}[-\text{len}(\textcolor{darkred}{R_K}):]$
    \State $\textcolor{darkred}{t_O} \gets \textcolor{darkred}{A_{\text{quant}}} \mathbf{q^{-1}(}\textcolor{darkblue}{Q_V}\mathbf{)} + \textcolor{darkred}{A_{\text{unquant}}} \textcolor{darkred}{R_V}$
    
    \State $\text{KV Cache} \gets \textcolor{darkblue}{Q_K}, \textcolor{darkblue}{Q_V}, \textcolor{darkred}{R_K}, \textcolor{darkred}{R_V}$
    \State \Return $\textcolor{darkred}{t_O}$
\EndProcedure
\end{algorithmic}
\end{algorithm}

\section{Experiments}
\label{sec:eval}

We conduct experiments to evaluate the effectiveness of \name in improving accuracy preserving and inference performance.

\subsection{Evaluation Methodology}
\textbf{Models.} We compare \name against state-of-the-art public LLMs, including LLaMA2-7B-chat, LLaMA2-13B-chat~\cite{llama2} and Mistral-7B-Instruct-v0.2 ~\cite{mistral}. 

\definecolor{lightbrown}{rgb}{0.85, 0.65, 0.5}
\definecolor{darkerbrown}{rgb}{0.7, 0.45, 0.25}
\begin{table*}[!ht]

\fontsize{18}{24}\selectfont
\setlength{\tabcolsep}{5pt}
\centering

\caption{Performance evaluation of \name on various models in a range of benchmarks in LongBench. Rows marked in \textcolor{darkerbrown}{brown} have a similar KV cache size, while KIVI and the full model use a larger KV cache.}
\label{tab1:longbench}
\begin{threeparttable}

\scalebox{0.31}{
\begin{tabular}{l|lcccccccccccccc}
\specialrule{1pt}{2pt}{2pt}

\multirow{4}{*}{~~~~\textbf{Models}} &\multirow{4}{*}{~~~~\textbf{Methods}} & \multicolumn{2}{c}{\textbf{Single-Doc QA}} & \multicolumn{2}{c}{\textbf{Synthetic}}& \multicolumn{2}{c}{\textbf{Code}}& \multicolumn{2}{c}{\textbf{Multi-Doc QA}}& \multicolumn{2}{c}{\textbf{Summarization}} & \multicolumn{3}{c}{\textbf{Few-Shot Learning}}  \\
\cmidrule(lr){3-4}\cmidrule(lr){5-6}\cmidrule(lr){7-8}\cmidrule(lr){9-10}\cmidrule(lr){11-12}\cmidrule(lr){13-15}

&& \rotatebox[origin=c]{30}{Qasper} & \rotatebox[origin=c]{30}{MultifieldQA} & \rotatebox[origin=c]{30}{Passage Ret.} & \rotatebox[origin=c]{30}{Passage Ct.} & \rotatebox[origin=c]{30}{LCC} & \rotatebox[origin=c]{30}{RepoBench-P} & \rotatebox[origin=c]{30}{2WikiMQA} & \rotatebox[origin=c]{30}{HotpotQA} & \rotatebox[origin=c]{30}{Gov Report} & \rotatebox[origin=c]{30}{Multi News} & \rotatebox[origin=c]{30}{TREC} & \rotatebox[origin=c]{30}{SamSum} & \rotatebox[origin=c]{30}{TriviaQA} & \rotatebox[origin=c]{30}{Average} \\ 

\specialrule{1pt}{2pt}{10pt}

\multirow{6}{*}{\rotatebox[origin=c]{0}{\fontsize{20}{100} \textbf{LLaMA2-7B-chat}}}
& \cellcolor{teal!20} Full Model & \cellcolor{teal!20} 22.78 & \cellcolor{teal!20} 33.59 & \cellcolor{teal!20} 8.44 & \cellcolor{teal!20} 4.75 & \cellcolor{teal!20} 59.56 & \cellcolor{teal!20} 48.07 & \cellcolor{teal!20} 22.35 & \cellcolor{teal!20} 24.88 & \cellcolor{teal!20} 24.99 & \cellcolor{teal!20} 23.60 & \cellcolor{teal!20} 59.67 & \cellcolor{teal!20} 39.38 & \cellcolor{teal!20} 85.38 & \cellcolor{teal!20} 35.19 \\

& KIVI &  22.45 & 33.32 & 11.33 & 4.25 & 59.05 & 47.96 & 21.88 & 23.88 & 24.46 & 22.86 & 59.67 & 38.74 & 84.80 & 34.97 \\
& \cellcolor{darkerbrown!20} \ho (15\%) & \cellcolor{darkerbrown!20} 16.98 & \cellcolor{darkerbrown!20} 29.72 & \cellcolor{darkerbrown!20} 11.00 & \cellcolor{darkerbrown!20} 4.55 & \cellcolor{darkerbrown!20} 56.87 & \cellcolor{darkerbrown!20} 48.25 & \cellcolor{darkerbrown!20} 19.92 & \cellcolor{darkerbrown!20} 24.58 & \cellcolor{darkerbrown!20} 22.19 & \cellcolor{darkerbrown!20} 22.16 & \cellcolor{darkerbrown!20} 57.33 & \cellcolor{darkerbrown!20} 37.80 & \cellcolor{darkerbrown!20} 84.02 & \cellcolor{darkerbrown!20} 33.49 \\
& \cellcolor{darkerbrown!20} \snap (15\%) & \cellcolor{darkerbrown!20} 17.41 & \cellcolor{darkerbrown!20} 34.53 & \cellcolor{darkerbrown!20} 8.67 & \cellcolor{darkerbrown!20} 3.59 & \cellcolor{darkerbrown!20} 58.48 & \cellcolor{darkerbrown!20} 47.52 & \cellcolor{darkerbrown!20} 21.00 & \cellcolor{darkerbrown!20} 24.91 & \cellcolor{darkerbrown!20} 19.04 & \cellcolor{darkerbrown!20} 19.74 & \cellcolor{darkerbrown!20} 59.33 & \cellcolor{darkerbrown!20} 37.92 & \cellcolor{darkerbrown!20} 84.72 & \cellcolor{darkerbrown!20} 33.60 \\
& \cellcolor{darkerbrown!20} Q-Hitter (59\%) & \cellcolor{darkerbrown!20} 17.43 & \cellcolor{darkerbrown!20} 30.08 & \cellcolor{darkerbrown!20} 9.00 & \cellcolor{darkerbrown!20} 4.13 & \cellcolor{darkerbrown!20} 56.84 & \cellcolor{darkerbrown!20} 45.18 & \cellcolor{darkerbrown!20} 17.66 & \cellcolor{darkerbrown!20} 22.57 & \cellcolor{darkerbrown!20} 22.83 & \cellcolor{darkerbrown!20} 22.48 & \cellcolor{darkerbrown!20} 59.67 & \cellcolor{darkerbrown!20} 38.46 & \cellcolor{darkerbrown!20} 82.76 & \cellcolor{darkerbrown!20} 33.01 \\

& \cellcolor{darkerbrown!20} \name & \cellcolor{darkerbrown!20} 21.01 & \cellcolor{darkerbrown!20} 29.23 & \cellcolor{darkerbrown!20} 10.00 & \cellcolor{darkerbrown!20} 3.82 & \cellcolor{darkerbrown!20} 58.38 & \cellcolor{darkerbrown!20} 47.99 & \cellcolor{darkerbrown!20} 20.91 & \cellcolor{darkerbrown!20} 22.97 & \cellcolor{darkerbrown!20} 23.45 & \cellcolor{darkerbrown!20} 22.54 & \cellcolor{darkerbrown!20} 59.00 & \cellcolor{darkerbrown!20} 37.94 & \cellcolor{darkerbrown!20} 80.95 & \cellcolor{darkerbrown!20} 33.71 \\
& \cellcolor{darkerbrown!20} \name Pyramid & \cellcolor{darkerbrown!20} 19.92 & \cellcolor{darkerbrown!20} 33.96 & \cellcolor{darkerbrown!20} 10.00 & \cellcolor{darkerbrown!20} 4.12 & \cellcolor{darkerbrown!20} 59.72 & \cellcolor{darkerbrown!20} 49.29 & \cellcolor{darkerbrown!20} 20.69 & \cellcolor{darkerbrown!20} 24.62 & \cellcolor{darkerbrown!20} 24.16 & \cellcolor{darkerbrown!20} 22.90 & \cellcolor{darkerbrown!20} 59.00 & \cellcolor{darkerbrown!20} 39.15 & \cellcolor{darkerbrown!20} 82.89 & \cellcolor{darkerbrown!20} 34.65 \\

\specialrule{1pt}{2pt}{10pt}

\multirow{6}{*}{\rotatebox[origin=c]{0}{\fontsize{20}{100} \textbf{LLaMA2-13B-chat}}}
& \cellcolor{teal!20} Full Model & \cellcolor{teal!20} 13.72 & \cellcolor{teal!20} 28.11 & \cellcolor{teal!20} 20.67 & \cellcolor{teal!20} 5.58 & \cellcolor{teal!20} 49.97 & \cellcolor{teal!20} 47.18 & \cellcolor{teal!20} 12.13 & \cellcolor{teal!20} 15.14 & \cellcolor{teal!20} 26.29 & \cellcolor{teal!20} 23.52 & \cellcolor{teal!20} 64.00 & \cellcolor{teal!20} 40.39 & \cellcolor{teal!20} 86.52 & \cellcolor{teal!20} 33.32 \\
& KIVI &  13.56 & 28.16 & 17.33 & 5.05 & 49.21 & 47.18 & 12.80 & 15.27 & 25.24 & 23.07 & 64.33 & 40.24 & 87.07 & 32.96 \\
& \cellcolor{darkerbrown!20} \ho (15\%) & \cellcolor{darkerbrown!20} 11.94 & \cellcolor{darkerbrown!20} 25.13 & \cellcolor{darkerbrown!20} 15.67 & \cellcolor{darkerbrown!20} 4.61 & \cellcolor{darkerbrown!20} 48.18 & \cellcolor{darkerbrown!20} 44.29 & \cellcolor{darkerbrown!20} 13.04 & \cellcolor{darkerbrown!20} 14.52 & \cellcolor{darkerbrown!20} 23.15 & \cellcolor{darkerbrown!20} 22.12 & \cellcolor{darkerbrown!20} 59.67 & \cellcolor{darkerbrown!20} 39.66 & \cellcolor{darkerbrown!20} 83.70 & \cellcolor{darkerbrown!20} 31.2 \\
& \cellcolor{darkerbrown!20} SnapKV (15\%) & \cellcolor{darkerbrown!20} 12.11 & \cellcolor{darkerbrown!20} 27.09 & \cellcolor{darkerbrown!20} 22.00 & \cellcolor{darkerbrown!20} 5.18 & \cellcolor{darkerbrown!20} 49.52 & \cellcolor{darkerbrown!20} 45.44 & \cellcolor{darkerbrown!20} 14.10 & \cellcolor{darkerbrown!20} 14.40 & \cellcolor{darkerbrown!20} 20.06 & \cellcolor{darkerbrown!20} 20.75 & \cellcolor{darkerbrown!20} 62.33 & \cellcolor{darkerbrown!20} 39.25 & \cellcolor{darkerbrown!20} 85.86 & \cellcolor{darkerbrown!20} 32.16 \\
& \cellcolor{darkerbrown!20} \name & \cellcolor{darkerbrown!20} 11.24 & \cellcolor{darkerbrown!20} 25.13 & \cellcolor{darkerbrown!20} 15.00 & \cellcolor{darkerbrown!20} 3.62 & \cellcolor{darkerbrown!20} 48.43 & \cellcolor{darkerbrown!20} 46.10 & \cellcolor{darkerbrown!20} 12.74 & \cellcolor{darkerbrown!20} 16.16 & \cellcolor{darkerbrown!20} 24.26 & \cellcolor{darkerbrown!20} 22.84 & \cellcolor{darkerbrown!20} 63.33 & \cellcolor{darkerbrown!20} 40.79 & \cellcolor{darkerbrown!20} 84.33 & \cellcolor{darkerbrown!20} 31.84 \\
& \cellcolor{darkerbrown!20} \name Pyramid & \cellcolor{darkerbrown!20} 12.79 & \cellcolor{darkerbrown!20} 27.32 & \cellcolor{darkerbrown!20} 17.00 & \cellcolor{darkerbrown!20} 2.79 & \cellcolor{darkerbrown!20} 48.94 & \cellcolor{darkerbrown!20} 46.25 & \cellcolor{darkerbrown!20} 12.66 & \cellcolor{darkerbrown!20} 15.47 & \cellcolor{darkerbrown!20} 25.06 & \cellcolor{darkerbrown!20} 23.14 & \cellcolor{darkerbrown!20} 63.67 & \cellcolor{darkerbrown!20} 40.35 & \cellcolor{darkerbrown!20} 85.33 & \cellcolor{darkerbrown!20} 32.37 \\

\specialrule{1pt}{2pt}{10pt}

\multirow{6}{*}{\rotatebox[origin=c]{0}{\fontsize{20}{100} \textbf{Mistral7B-instruct}}}
& \cellcolor{teal!20} Full Model & \cellcolor{teal!20} 25.79 & \cellcolor{teal!20} 47.97 & \cellcolor{teal!20} 50.83 & \cellcolor{teal!20} 2.98 & \cellcolor{teal!20} 50.69 & \cellcolor{teal!20} 47.22 & \cellcolor{teal!20} 27.44 & \cellcolor{teal!20} 36.44 & \cellcolor{teal!20} 31.84 & \cellcolor{teal!20} 25.82 & \cellcolor{teal!20} 62.67 & \cellcolor{teal!20} 40.49 & \cellcolor{teal!20} 86.29 & \cellcolor{teal!20} 41.2 \\
& KIVI &  25.13 & 46.30 & 50.75 & 3.02 & 51.16 & 46.81 & 26.39 & 35.11 & 31.23 & 25.36 & 62.33 & 40.12 & 86.31 & 40.77 \\
& \cellcolor{darkerbrown!20} \ho (15\%) & \cellcolor{darkerbrown!20} 20.20 & \cellcolor{darkerbrown!20} 42.55 & \cellcolor{darkerbrown!20} 42.84 & \cellcolor{darkerbrown!20} 3.00 & \cellcolor{darkerbrown!20} 49.66 & \cellcolor{darkerbrown!20} 45.95 & \cellcolor{darkerbrown!20} 24.27 & \cellcolor{darkerbrown!20} 33.04 & \cellcolor{darkerbrown!20} 27.43 & \cellcolor{darkerbrown!20} 24.33 & \cellcolor{darkerbrown!20} 60.33 & \cellcolor{darkerbrown!20} 40.45 & \cellcolor{darkerbrown!20} 86.20 & \cellcolor{darkerbrown!20} 38.4 \\
& \cellcolor{darkerbrown!20} SnapKV (15\%) & \cellcolor{darkerbrown!20} 24.14 & \cellcolor{darkerbrown!20} 48.32 & \cellcolor{darkerbrown!20} 50.23 & \cellcolor{darkerbrown!20} 3.04 & \cellcolor{darkerbrown!20} 50.39 & \cellcolor{darkerbrown!20} 45.76 & \cellcolor{darkerbrown!20} 25.76 & \cellcolor{darkerbrown!20} 34.55 & \cellcolor{darkerbrown!20} 25.10 & \cellcolor{darkerbrown!20} 22.77 & \cellcolor{darkerbrown!20} 61.67 & \cellcolor{darkerbrown!20} 40.12 & \cellcolor{darkerbrown!20} 86.90 & \cellcolor{darkerbrown!20} 39.90 \\
& \cellcolor{darkerbrown!20} \name & \cellcolor{darkerbrown!20} 22.94 & \cellcolor{darkerbrown!20} 45.80 & \cellcolor{darkerbrown!20} 49.47 & \cellcolor{darkerbrown!20} 3.36 & \cellcolor{darkerbrown!20} 49.78 & \cellcolor{darkerbrown!20} 45.56 & \cellcolor{darkerbrown!20} 24.27 & \cellcolor{darkerbrown!20} 33.84 & \cellcolor{darkerbrown!20} 29.73 & \cellcolor{darkerbrown!20} 25.22 & \cellcolor{darkerbrown!20} 61.67 & \cellcolor{darkerbrown!20} 39.96 & \cellcolor{darkerbrown!20} 86.36 & \cellcolor{darkerbrown!20} 39.84 \\
& \cellcolor{darkerbrown!20} \name Pyramid & \cellcolor{darkerbrown!20} 23.10 & \cellcolor{darkerbrown!20} 45.91 & \cellcolor{darkerbrown!20} 48.88 & \cellcolor{darkerbrown!20} 3.24 & \cellcolor{darkerbrown!20} 50.34 & \cellcolor{darkerbrown!20} 45.41 & \cellcolor{darkerbrown!20} 25.18 & \cellcolor{darkerbrown!20} 34.04 & \cellcolor{darkerbrown!20} 29.69 & \cellcolor{darkerbrown!20} 25.32 & \cellcolor{darkerbrown!20} 61.67 & \cellcolor{darkerbrown!20} 40.17 & \cellcolor{darkerbrown!20} 86.63 & \cellcolor{darkerbrown!20} 39.97 \\

\specialrule{1pt}{2pt}{10pt}

\end{tabular}
}
\end{threeparttable}
\end{table*}

\noindent
\textbf{Datasets.} We choose LongBench for evaluation~\cite{longbench},
which has been widely adopted in state-of-the-art works~\cite{kivi,kvquant,li2024snapkv}.
Additional details on the datasets used can be found in the Appendix~\ref{appenxi:dataset}.

\noindent
\textbf{Baselines.}
We compare \name with the following baselines: adaptive KV (\ho, \snap ~\cite{h2o, li2024snapkv}), INT2 quantized KV (KIVI~\cite{kivi}), adaptive + quantized KV (Q-hitter ~\cite{q-hitter}), and FullKV. Q-Hitter combines \ho with INT4 quantization, providing a strong baseline for \name.

\noindent
\textbf{Hyperparameters.}
We use a 50\% cache budget with \name, with 25\% heavy hitter budget and 25\% the recent window budget. 
The group size during token/channel-wise quantization is set to 16, i.e. 16 values along the token/channel axis share quantization zero point and scale. A residual length of $n_r = 128$ is used for both \name and KIVI. 
The maximum prompt length is 4096 for all models with the first and last 2048 tokens taken for a longer prompt. The maximum generation length is dataset-specific. No task has a generation length of more than $512$ tokens.
Please see Appendix \ref{sec::detail} for other evaluation details.

\noindent
\textbf{Hardware.}
We conducted experiments on NVIDIA 4$\times$A100-40GB, 4$\times$A40-46GB and 4$\times$GH200-120GB GPUs.

\subsection{Enhancing KV Cache Compression Accuracy in Long Context Inference}



To make a fair comparison, we compare all methods with adaptive KV policies (\ho, \snap, Q-Hitter, and \name)  under a similar KV cache size (Appendix \ref{sec:kv_mem_formulas}).
Given a prompt length of $4096$ and generation length of $512$, the KV cache size for \name is $0.33 \textrm{ GB}$.
A cache budget of $\alpha=15\%$ results in a similar KV cache size for \ho. A cache budget of $\alpha = 59\%$ results in a similar KV cache size for Q-Hitter. We test two strategies of \name, namely \name and \name-Pyramid, to demonstrate the effectiveness of \name. \name follows a uniform cache allocation with $(25\%, 25\%)$ HH, RW budget per layer. \name-Pyramid uses $25\%$ RW budget per layer but the HH budget is distributed across layers as described in \sref{sec:layer_wise_disc}.

The results are shown in Table~\ref{tab1:longbench}. 
\name outperforms other state-of-the-art adaptive KV methods (\ho, \snap, Q-Hitter) for the same KV cache size.
For LLaMA2-7B-chat, \name-Pyramid achieves an average accuracy of 
$34.65$, 
obtaining 98.5\% of the full model accuracy $35.19$. 
\name is also able to maintain accuracy on LLaMA2-13B-chat and Mistral-7B, indicating that our approach generalizes well across datasets and model classes. While the full model and KIVI perform marginally better than \name, they have much larger KV cache memory consumption. The synergistic composition of 2-bit quantized KV and layer-wise adaptive KV delivers these improvements, and it also shows the promising aspect of using both quantization and adaptive KV in conjunction to reduce the high memory footprint of the KV cache.

\subsection{Setting A New Pareto Frontier}
\label{sec:perf_vs_kv_size}

With \ho, \snap, Q-Hitter, and \name the user can tune the cache budget, potentially improving performance at the cost of a larger KV cache. An ideal technique would maintain performance when lowering the cache budget.
We plot the performance of \name against the KV cache size. The size of the KV cache is computed using the KV memory consumption analysis in Appendix \ref{sec:kv_mem_formulas}.
To highlight interesting configurations, we mark the Pareto optimal front, which is the configuration that offers the smallest KV cache size for the highest performance.

\begin{figure}[ht!]
    \centering
    \includegraphics[width=\linewidth]{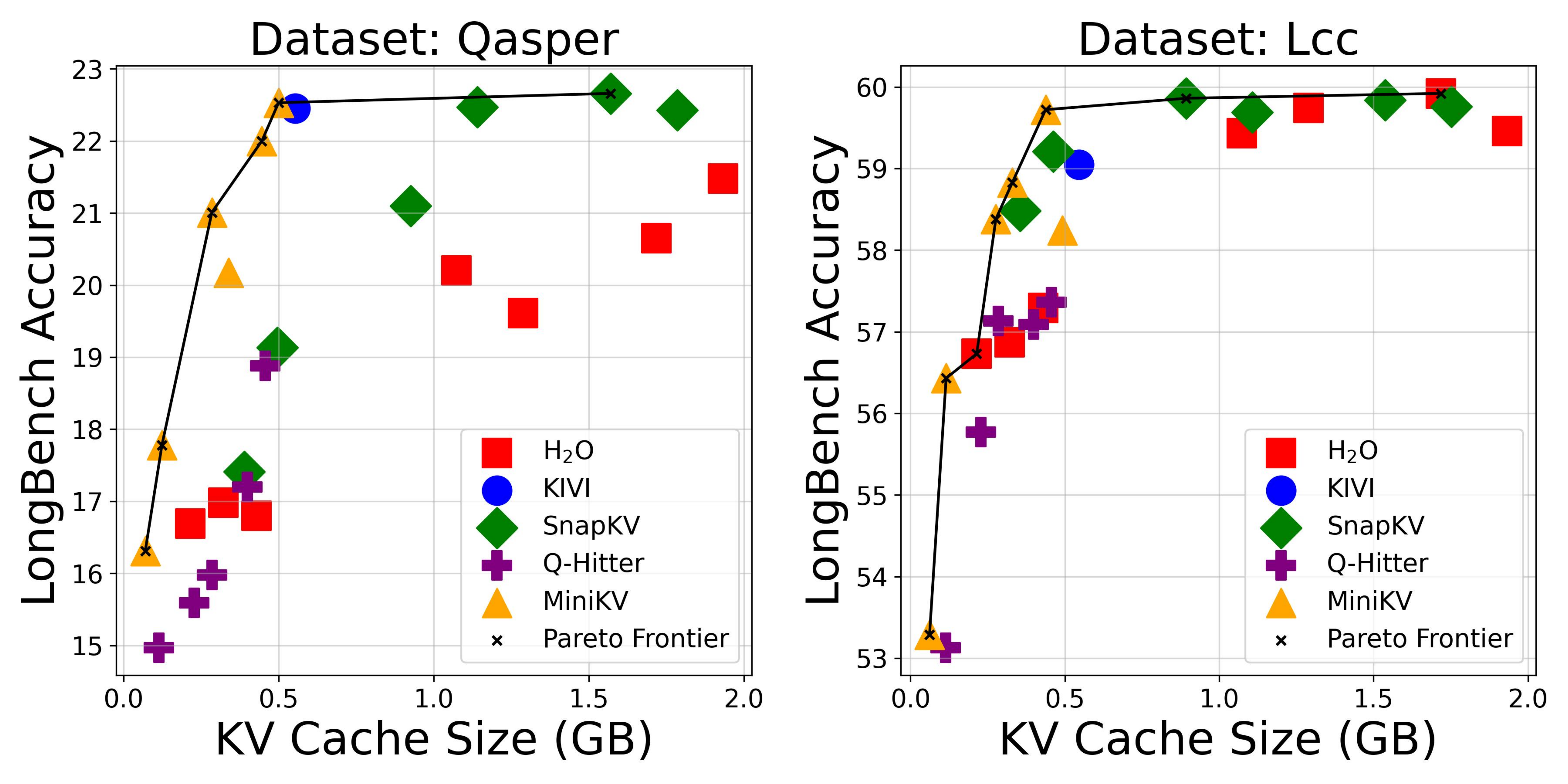}
    \caption{Algorithm Performance vs KV Cache Size: The Pareto frontier (the black curve) indicates the optimal compression strategy across a range of KV cache sizes. \name lies on the Pareto frontier across all 6 task categories.}
    \label{fig:perf_vs_kv_size}
\end{figure}

\fref{fig:perf_vs_kv_size} shows the performance vs KV cache size curve for two datasets (Qasper and Lcc), the remaining plots can be found in the Appendix \ref{appendix:perf_vs_kv_size}. \name achieves the pareto optimal compression strategy across all 6 major task categories on LongBench (single/multi-doc QA, LC understanding, code completion, summarization and few-shot learning). These results validate the effectiveness of \name with varying KV cache sizes. 

\subsection{Results on InfiniteBench}
We test the Llama3 herd of models~\cite{llama3} on selected datasets from the InfiniteBench benchmark~\cite{zhang2024infty} on GH200 GPUs. We compare \name with the uncompressed model baseline and the quantization-only baseline (KIVI). The results are shown in Table~\ref{tab1:infinitebench}.

\definecolor{lightbrown}{rgb}{0.85, 0.65, 0.5}
\definecolor{darkerbrown}{rgb}{0.7, 0.45, 0.25}
\begin{table*}[!ht]

\fontsize{18}{24}\selectfont
\setlength{\tabcolsep}{5pt}
\centering

\caption{Performance evaluation of \name on various models in a range of benchmarks in InfiniteBench. Rows marked in \textcolor{darkerbrown}{brown} have a similar KV cache size, while KIVI and the full model use a larger KV cache.}
\label{tab1:infinitebench}
\begin{threeparttable}

\scalebox{0.363}{
\begin{tabular}{l|lcccccccccc}
\specialrule{1pt}{2pt}{2pt}

\multirow{2}{*}{~~~~\textbf{Models}} & \multirow{2}{*}{~~~~\textbf{Methods}} & \multicolumn{8}{c}{\textbf{Benchmarks}} & \multirow{2}{*}{\textbf{Average}} \\
\cmidrule(lr){3-10}

& & \rotatebox[origin=c]{30}{LongBook Choice} & \rotatebox[origin=c]{30}{LongDialogue QA} & \rotatebox[origin=c]{30}{Code Debug} & \rotatebox[origin=c]{30}{Passkey} & \rotatebox[origin=c]{30}{Number String} & \rotatebox[origin=c]{30}{KV Retrieval} & \rotatebox[origin=c]{30}{LongBook QA} & \rotatebox[origin=c]{30}{Math Find} \\

\specialrule{1pt}{2pt}{10pt}

\multirow{4}{*}{\rotatebox[origin=c]{0}{\fontsize{20}{100} \textbf{Llama3-8b-instruct}}}
& \cellcolor{teal!20} Full Model & \cellcolor{teal!20}39.74 & \cellcolor{teal!20}12.00 & \cellcolor{teal!20}31.22 & \cellcolor{teal!20}3.39 & \cellcolor{teal!20}3.39 & \cellcolor{teal!20}1.00 & \cellcolor{teal!20}5.94 & \cellcolor{teal!20}12.00 & \cellcolor{teal!20}13.59 \\
& KIVI & 39.74 & 11.50 & 31.22 & 3.39 & 3.39 & 0.80 & 6.11 & 12.00 & 13.52 \\
& \cellcolor{darkerbrown!20} MiniKV & \cellcolor{darkerbrown!20}40.17 & \cellcolor{darkerbrown!20}12.00 & \cellcolor{darkerbrown!20}31.47 & \cellcolor{darkerbrown!20}3.39 & \cellcolor{darkerbrown!20}2.20 & \cellcolor{darkerbrown!20}0.00 & \cellcolor{darkerbrown!20}6.12 & \cellcolor{darkerbrown!20}12.00 & \cellcolor{darkerbrown!20}13.42 \\
& \cellcolor{darkerbrown!20} MiniKV Pyramid & \cellcolor{darkerbrown!20}39.74 & \cellcolor{darkerbrown!20}12.50 & \cellcolor{darkerbrown!20}31.22 & \cellcolor{darkerbrown!20}3.39 & \cellcolor{darkerbrown!20}2.37 & \cellcolor{darkerbrown!20}0.20 & \cellcolor{darkerbrown!20}6.10 & \cellcolor{darkerbrown!20}12.00 & \cellcolor{darkerbrown!20}13.44 \\

\specialrule{1pt}{2pt}{10pt}

\multirow{4}{*}{\rotatebox[origin=c]{0}{\fontsize{20}{100} \textbf{Llama3-3b-instruct}}}
& \cellcolor{teal!20} Full Model & \cellcolor{teal!20}31.88 & \cellcolor{teal!20}12.00 & \cellcolor{teal!20}26.40 & \cellcolor{teal!20}3.39 & \cellcolor{teal!20}3.05 & \cellcolor{teal!20}0.60 & \cellcolor{teal!20}9.66 & \cellcolor{teal!20}12.57 & \cellcolor{teal!20}12.44 \\
& KIVI & 33.62 & 15.50 & 26.40 & 3.39 & 3.22 & 0.20 & 9.20 & 7.71 & 12.41 \\
& \cellcolor{darkerbrown!20} MiniKV & \cellcolor{darkerbrown!20}33.62 & \cellcolor{darkerbrown!20}11.00 & \cellcolor{darkerbrown!20}26.65 & \cellcolor{darkerbrown!20}3.05 & \cellcolor{darkerbrown!20}2.03 & \cellcolor{darkerbrown!20}0.00 & \cellcolor{darkerbrown!20}8.63 & \cellcolor{darkerbrown!20}8.00 & \cellcolor{darkerbrown!20}11.62 \\
& \cellcolor{darkerbrown!20} MiniKV Pyramid & \cellcolor{darkerbrown!20}34.50 & \cellcolor{darkerbrown!20}11.50 & \cellcolor{darkerbrown!20}26.65 & \cellcolor{darkerbrown!20}3.39 & \cellcolor{darkerbrown!20}1.86 & \cellcolor{darkerbrown!20}0.00 & \cellcolor{darkerbrown!20}8.92 & \cellcolor{darkerbrown!20}9.14 & \cellcolor{darkerbrown!20}11.99 \\

\specialrule{1pt}{2pt}{10pt}

\multirow{4}{*}{\rotatebox[origin=c]{0}{\fontsize{20}{100} \textbf{Llama3-1b-instruct}}}
& \cellcolor{teal!20} Full Model & \cellcolor{teal!20}37.55 & \cellcolor{teal!20}9.50 & \cellcolor{teal!20}24.87 & \cellcolor{teal!20}3.39 & \cellcolor{teal!20}3.22 & \cellcolor{teal!20}0.00 & \cellcolor{teal!20}10.71 & \cellcolor{teal!20}14.57 & \cellcolor{teal!20}12.98 \\
& KIVI & 37.12 & 8.50 & 24.87 & 3.39 & 2.71 & 0.00 & 10.37 & 12.86 & 12.48 \\
& \cellcolor{darkerbrown!20} MiniKV & \cellcolor{darkerbrown!20}37.12 & \cellcolor{darkerbrown!20}9.00 & \cellcolor{darkerbrown!20}24.87 & \cellcolor{darkerbrown!20}2.88 & \cellcolor{darkerbrown!20}1.69 & \cellcolor{darkerbrown!20}0.00 & \cellcolor{darkerbrown!20}10.01 & \cellcolor{darkerbrown!20}14.86 & \cellcolor{darkerbrown!20}12.55 \\
& \cellcolor{darkerbrown!20} MiniKV Pyramid & \cellcolor{darkerbrown!20}37.12 & \cellcolor{darkerbrown!20}9.50 & \cellcolor{darkerbrown!20}24.37 & \cellcolor{darkerbrown!20}3.05 & \cellcolor{darkerbrown!20}1.69 & \cellcolor{darkerbrown!20}0.00 & \cellcolor{darkerbrown!20}10.15 & \cellcolor{darkerbrown!20}14.29 & \cellcolor{darkerbrown!20}12.52 \\

\specialrule{1pt}{2pt}{10pt}

\end{tabular}
}
\end{threeparttable}
\end{table*}

For the Llama3-8B-instruct model, \name achieves an average score of $13.44$, closely matching the full model and KIVI's scores of $13.52$ and $13.59$, respectively, while utilizing a significantly smaller KV cache.

\subsection{Results on GSM8K}
\begin{figure}[ht!]
    \centering
    \includegraphics[width=\linewidth]{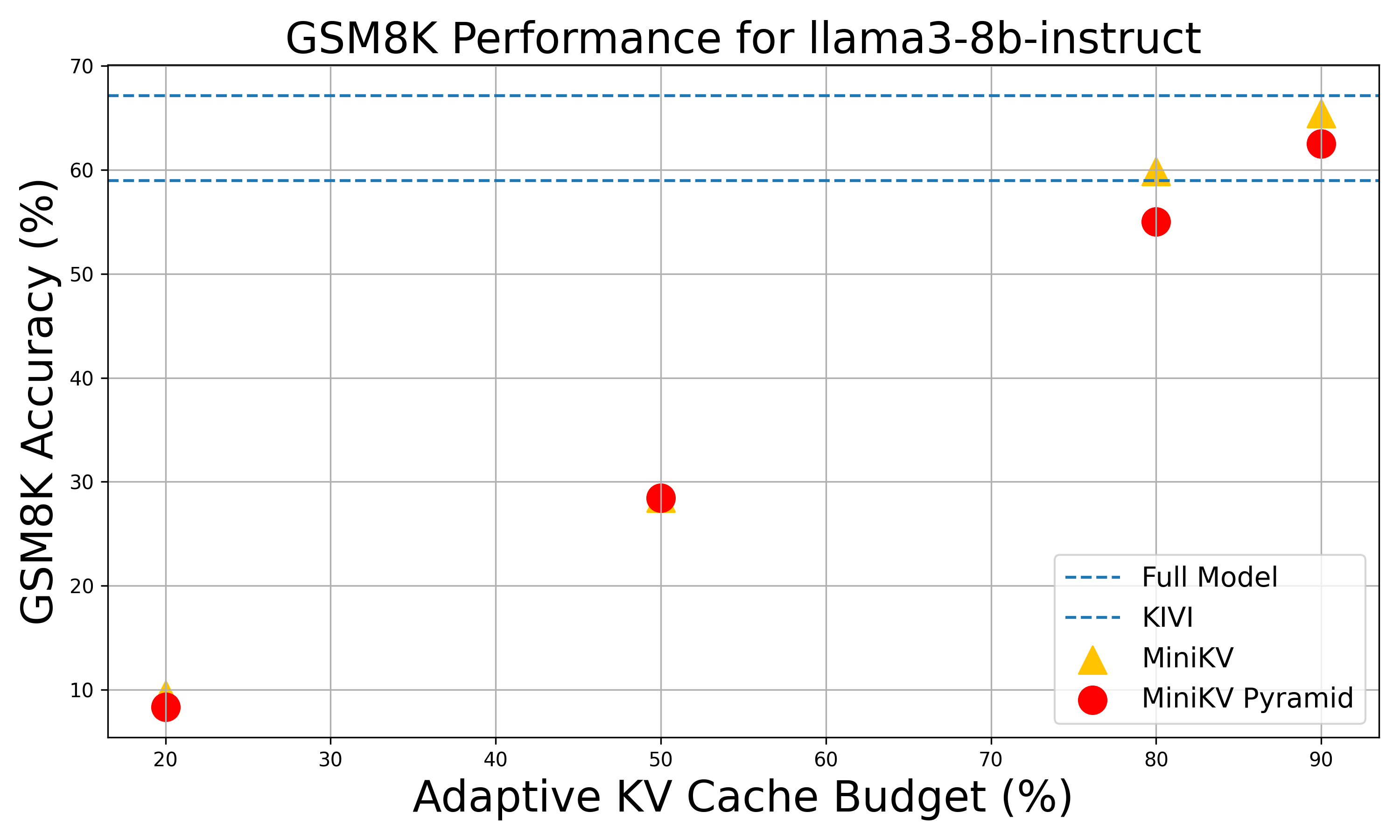}
    \caption{Performance on GSM8K: Since GSM8K is a reasoning-intensive task, \name requires a significant adaptive KV cache budget ($\sim90\%$) to match the performance of the full model.}
    \label{fig:gsm_lineplot}
\end{figure}

We evaluate the Llama3 model family~\cite{llama3} on the Platinum GSM8K dataset~\cite{vendrow2025large}, a reasoning-focused benchmark with short contexts ($\sim256$ tokens). Unlike long-context generation, where KV cache growth creates scalability issues, GSM8K poses a different challenge. Despite its compact inputs, the task requires strong KV state retention (90\% adaptive budget) for accurate reasoning, as shown in~\fref{fig:gsm_lineplot}.

\subsection{System Performance Results}
We evaluate the system performance of the LLaMA2-7B-chat model on a single NVIDIA A100 GPU with 40GB of memory.
We utilize FlashAttention kernels for KIVI and the Full Model while employing our customized kernel introduced in \sref{subsec: kernel} for \name. \ho and Q-Hitter do not support FlashAttention.

\begin{figure}[ht!]
    \centering
    \includegraphics[width=0.47\textwidth]{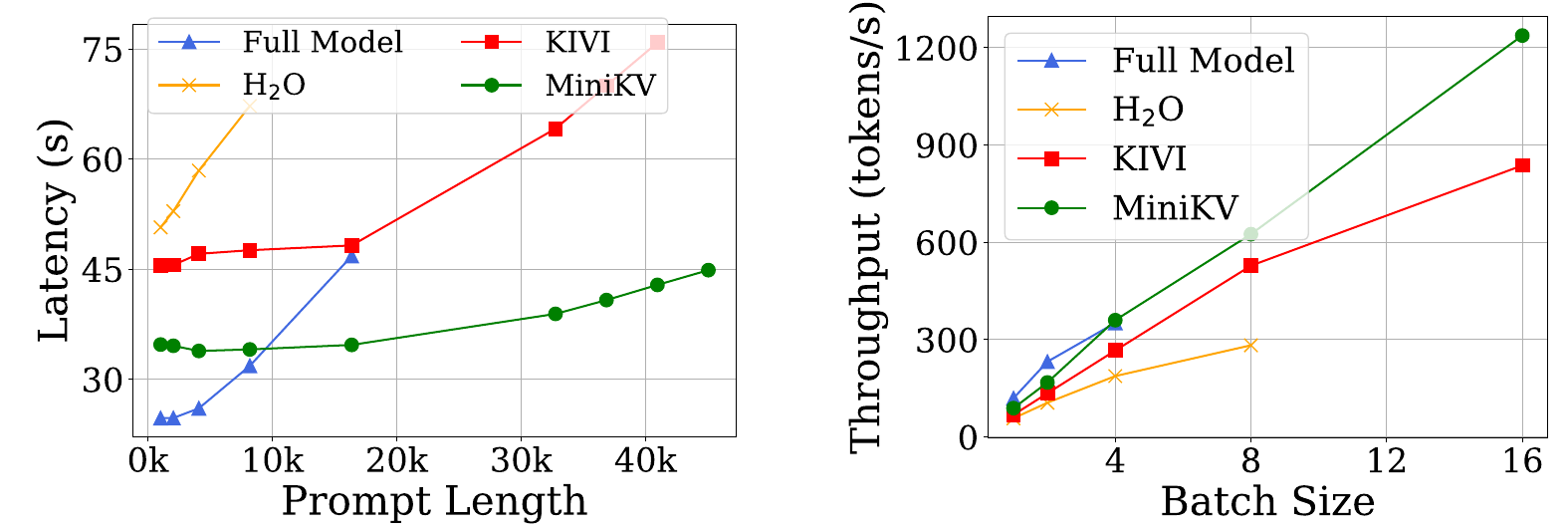}
    \caption{Left: Latency (s) for batch size = 1 and generation length = 1024. Right: Throughput (tokens/s) for prompt length = 2048 and generation length = 1024. }
    \label{fig:lat_thru}
\end{figure}

\noindent \textbf{Speeding up end-to-end latency.}
LLM inference is predominantly constrained by the memory bandwidth required to retrieve the model states. \name reduces latency through a compression and system co-design approach, which reduces the number of KV pairs loaded for each next-token prediction by revising 2-bit KV quantization combined with adaptive KV policies, while at the same time maintaining hardware friendly execution using high-performance memory-efficient kernels compatible with system optimizations such as FlashAttention. As a result, as shown in \fref{fig:lat_thru} (left), \name has a lower latency than its baselines, especially in long sequences (e.g., $>$10k). We include a detailed latency breakdown analysis in Appendix~\ref{appendix:e2e-breakdown}. 




\noindent \textbf{Achieving high throughput.} As shown in \fref{fig:lat_thru} (right), \name outperforms all its baselines in throughput, measured as the number of tokens processed per second, due to its lower latency and ability to support larger batch sizes and longer sequence lengths.


\begin{figure}[ht!]
    \centering 
    \begin{subfigure}[b]{0.23\textwidth} 
        \includegraphics[width=\textwidth]{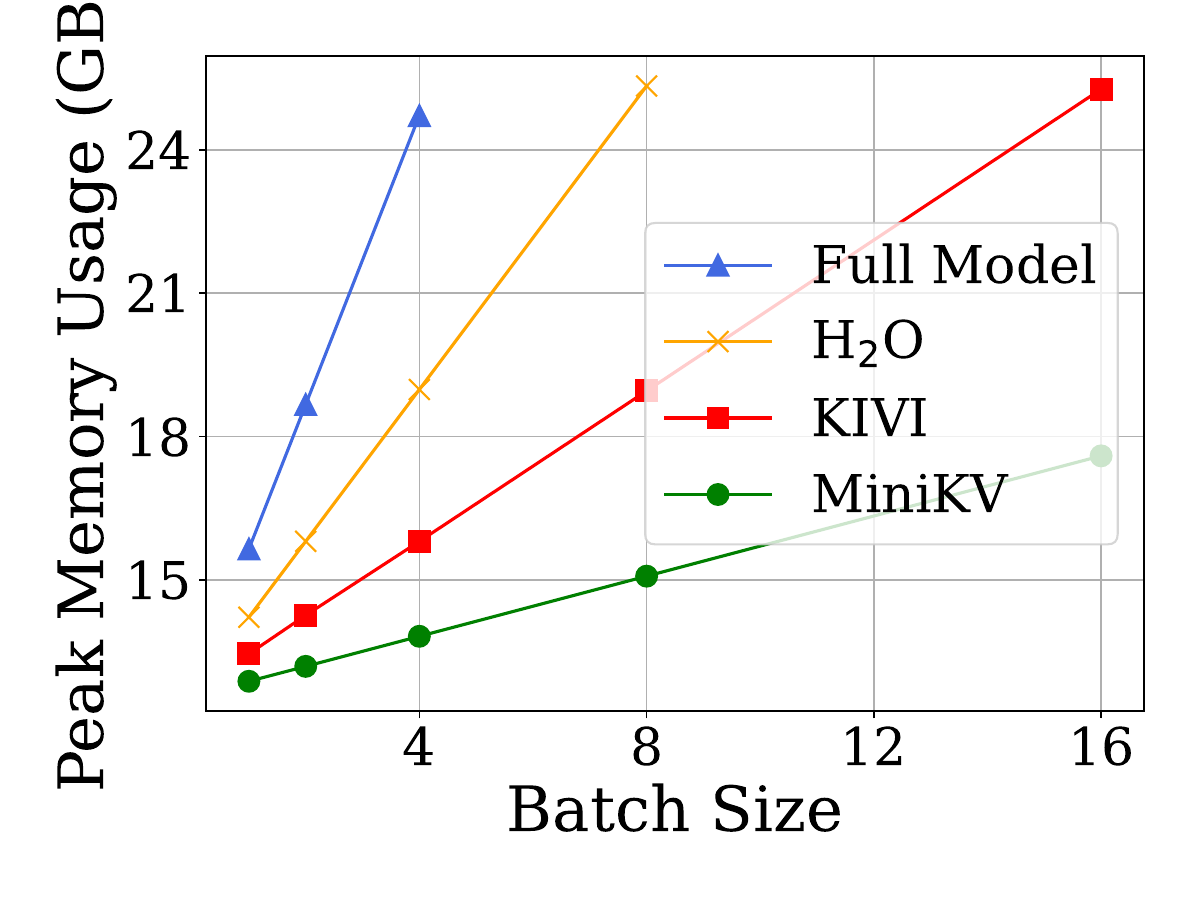}
    \end{subfigure}
    \begin{subfigure}[b]{0.23\textwidth} 
        \includegraphics[width=\textwidth]{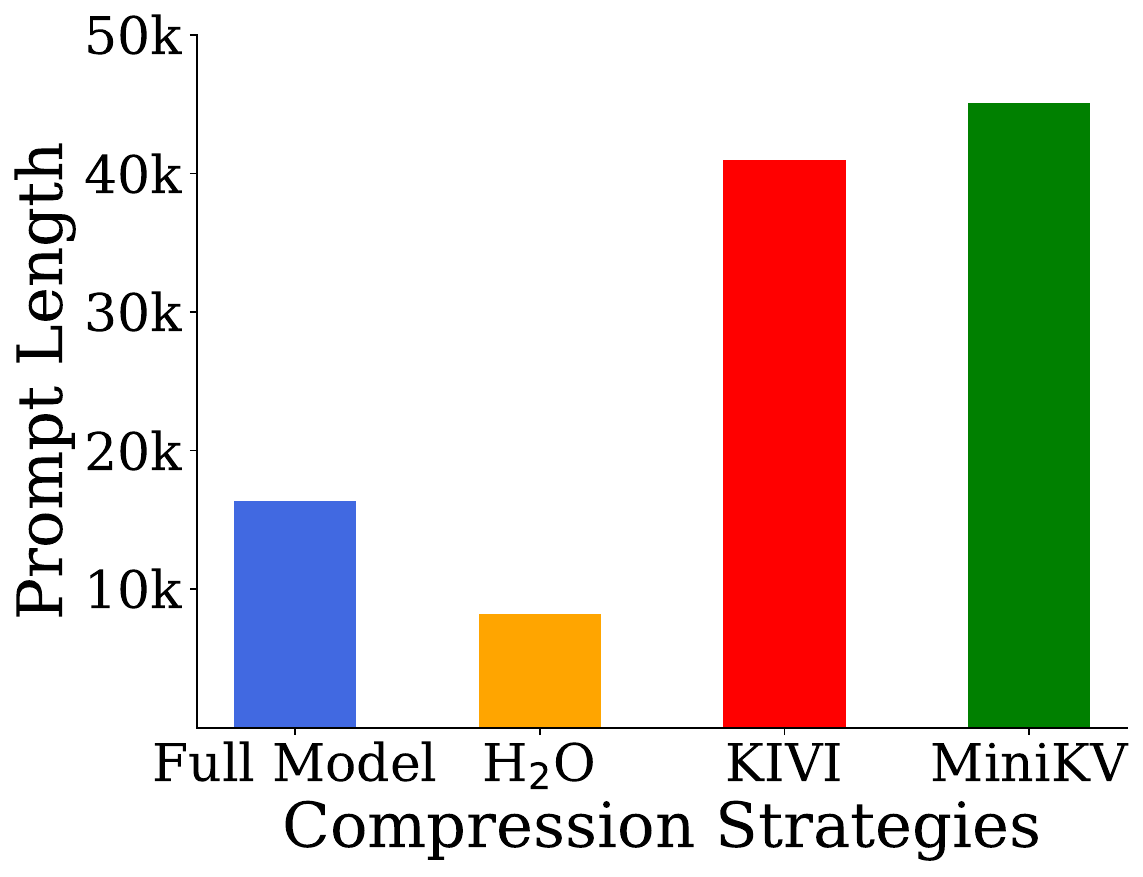}
    \end{subfigure}
    \caption{Left: Peak memory usage (GB) vs batch size for prompt = 2048 and generation length = 1024. Right: Maximum prompt length supported by \name and its baselines for batch size = 1.}
    \label{fig:mem_max_seq}
\end{figure}

\noindent \textbf{Effectively reducing peak memory usage.}
We benchmark peak memory usage, i.e., the maximum memory occupied by all model tensors during the generation. The memory savings achieved by KV cache compression can be rendered ineffective if peak memory usage exceeds the total GPU memory.
We evaluate the impact of batch size and prompt length on peak memory usage in \fref{fig:mem_max_seq} (left).
\name demonstrates the lowest peak memory consumption compared to its baselines. \ho goes out-of-memory at batch size 16 as it materializes the intermediate attention score matrix while KIVI maintains the full KV cache and therefore has a higher memory consumption.


\noindent \textbf{Enhancing maximum processable prompt.}
\name's lower memory consumption becomes more apparent with longer prompt lengths. \fref{fig:mem_max_seq} (right) shows that \name can process prompts 10\% longer than its strongest baseline KIVI. Additionally, \name's selective flash-attention kernel allows significantly longer sequence lengths when compared to \ho.

\noindent \textbf{Micro-benching on MiniKV's kernel.}
Table \ref{tab:memory-usage} and \ref{tab:latency} shows \name's attention kernel outperforms the standard attention implementation used in \ho. Unlike the standard attention mechanism, \name's memory footprint scales linearly with sequence length, allowing for much longer prompts. Furthermore, \name’s kernel offers significantly reduced latency, enabling faster processing.

\label{appendix:kernel-micro}
\begin{table}[ht]
\small
  \centering
  \renewcommand{\arraystretch}{1.4} 
  \begin{tabular}{|l|c|c|c|c|}
    \hline
    & \textbf{1024} & \textbf{2048} & \textbf{4096} & \textbf{8192} \\[0.5pt] \hline
    MiniKV      & 0.25  & 0.50  & 1.00   & 2.01 \\[0.5pt] \hline
    Standard & 1.25  & 4.51  & 17.01  & OOM  \\[0.5pt] \hline
  \end{tabular}
  \caption{Memory usage (GB) comparison between \name's kernel and the standard attention operator on different input sequence lengths.}
  \label{tab:memory-usage}
\end{table}
\vspace{0.5cm} 
\begin{table}[ht]
  \small
  \centering
  \renewcommand{\arraystretch}{1.4} 
  \begin{tabular}{|l|c|c|c|c|}
    \hline
    & \textbf{1024} & \textbf{2048} & \textbf{4096} & \textbf{8192} \\[0.5pt] \hline
    MiniKV      & 5.21   & 14.85  & 48.04   & 187.83 \\[0.5pt] \hline
    Standard & 10.46  & 40.18  & 130.51  & OOM    \\[0.5pt] \hline
  \end{tabular}
  \caption{Latency (ms) comparison between \name's kernel and the standard attention operator on different input sequence lengths.}
  \label{tab:latency}
\end{table}

MiniKV’s kernel is used only during the prefill phase to avoid being bottlenecked by the quadratic dependence on sequence length. While the kernel slows down the execution time of the prefill phase from 0.118 ms (using FlashAttention) to 0.622 ms (using \name's kernel) in LLama2-7B, it significantly reduces the memory consumption from 1.25 GB (using standard attention computation) to 0.25 GB (using \name's kernel) for sequence length 1024, effectively enabling longer sequence inference without running out-of-memory.

\section{Conclusion}
\label{sec:conclusion}

In this work, we revisit KV cache optimization via compression and system co-design to accelerate the inference of LLM. Our empirical analysis indicates that it is challenging to directly compose state-of-the-art 2-bit quantized KV with existing adaptive KV policies while preserving both accuracy and system efficiency on long context tasks under a high compression ratio. To address this issue, we develop \name to bridge the gap between ultra low-bit KV  quantization and adaptive policies, as well as the gap between the compression algorithm and hardware. Evaluation on a wide range of datasets and models shows that \name preserves long context accuracy while significantly improving the efficiency of LLM inference.

\cleardoublepage
\section{Limitations}

\name is promising in optimizing the KV cache. However, we identify several limitations and opportunities that can become future avenues of research to achieve an even higher compression ratio and generalizable compression.

\begin{enumerate}

    \item \textbf{Combination with model optimizations.} While we mainly focus on KV cache optimization (which provides significant benefits on its own), \name can also be combined with other optimization techniques, such as model compression~\cite{gptq,smoothquant}. This would further improve the computational and memory efficiency of LLMs.
    
    \item \textbf{Extensible design.} While we use \ho and KIVI as an example, our approach is compatible with other KV optimization techniques, such as StreamingLLM~\cite{attention-sink} and KVQuant~\cite{kvquant}.
    Given that \name combines \ho and KIVI, we also explored the possibility of combining \snap and KIVI.
    This combination should be viable in theory, as it involves only changing the eviction strategy during the prefill phase. However, we find that doing so leads to a severe drop in performance, with LongBench scores dropping from 35 to 32 points. Further experiments show that the tokens retained by \snap tend to be more sensitive to 2-bit quantization than those selected by \ho.
    
    This highlights the need for a more robust and generalizable approach to combining eviction and quantization, and a framework to determine when such combinations are effective.
\end{enumerate}

\section*{Acknowledgments}

We sincerely appreciate the insightful feedback from the anonymous reviewers. This research was supported by the National Science Foundation (NSF) under Grant No. 2441601. The work utilized the DeltaAI system at the National Center for Supercomputing Applications (NCSA) through allocation CIS240055 from the Advanced Cyberinfrastructure Coordination Ecosystem: Services \& Support (ACCESS) program, which is supported by National Science Foundation grants \#2138259, \#2138286, \#2138307, \#2137603, and \#2138296. The Delta advanced computing resource is a collaborative effort between the University of Illinois Urbana-Champaign and NCSA, supported by the NSF (award OAC 2005572) and the State of Illinois. This work also utilized the Illinois Campus Cluster and NCSA NFI Hydro cluster, both supported by the University of Illinois Urbana-Champaign and the University of Illinois System.

\bibliography{reference}

\appendix
\clearpage
\newpage

\appendix

\section{Formal Problem Formulation}
\label{sec:problem}

We introduce a general formulation of the co-compression of the KV cache via quantization and selection. For a given LLM $\Phi$ with $H$ layers, we denote its key states and value states at a layer $h$ as $\mathcal{K}_h\in \mathbb{R}^{n\times d}$ and  $\mathcal{V}_h\in \mathbb{R}^{n\times d}$, respectively. Let $Q_{h}\in \mathbb{R}^{1\times d}$ denote the query state. Then, the output $\mathcal{O}_{h\ }$ for each attention head of $\Phi$ is:
\begin{equation}
\label{eqn:attn}
\mathcal{O}_{h\ }=\mathcal{A}_{h}\mathcal{V}_{h},\  \mathcal{A}_{h}=softmax\left( \frac{Q_{h}\mathcal{K}_{h}^{T}}{\sqrt{d}} \right)
\end{equation}
Then the co-compression problem can be formulated as:

\noindent \textbf{Definition 2.1} (KV Cache Co-Compression Problem, informal). 

\noindent \emph{$\forall $ $\mathcal{K}_h$ and $\mathcal{V}_h$, where $h \in \{0,1,..,H-1\}$, find the quantizer $\mathcal{Q}_b[\cdot]$ with $b$ quantization bits, the selection policy $\mathcal{S}_h[\cdot]$ with $C$ selective KV cache size, such that $|\mathcal{O}_{h}-\mathcal{O}_{h}^{*}| \leq \epsilon$, where $\mathcal{O}_{h}^{*}$ represents the output for each attention head of $\Phi$ with $\mathcal{S}_h[\cdot]$ and $\mathcal{Q}_b[\cdot]$, and $\epsilon$ is an acceptable small positive value.
}

\section{Comparison of \name with Alternative Methods}
\label{appendix:copmarison}

We provide a detailed summary of the comparison between \name and previous approaches in Table~\ref{tab:related-work}. 

\begin{table*}[!ht]
\centering
\caption{Comparison with previous KV cache optimization methods for LLM inference.}
\label{tab:related-work}
\footnotesize
\begin{tabular}{|l|l|l|l|l|}
\hline
Approach       & Eviction-based KV & Quantization & Training-free & LongBench \\ \hline
AttentionSink~\cite{attention-sink} & \Checkmark           &              & \Checkmark              &              \\ \hline
FastGen~\cite{fastgen}     &  \Checkmark           &        &  \Checkmark      &             \\ \hline
ScissorHands~\cite{scissorhands}   & \Checkmark           &  4-bit            & \Checkmark              &              \\ \hline
H2O~\cite{h2o}            & \Checkmark           &  4-bit            & \Checkmark              &              \\ \hline
FlexGen~\cite{flexgen}        &             & 4-bit        & \Checkmark              &              \\ \hline
LLM-QAT~\cite{llm-qat}        &             & 4-bit        &                &              \\ \hline
Q-Hitter~\cite{q-hitter}     &  \Checkmark           & 4-bit       & \Checkmark              &             \\ \hline
KVQuant~\cite{kvquant}        &             & 4-bit        & \Checkmark              &   \Checkmark        \\ \hline
KIVI~\cite{kivi}           &             & 2-bit           & \Checkmark              & \Checkmark            \\ \hline
\name         & \Checkmark           & 2-bit        & \Checkmark            & \Checkmark            \\ \hline
\end{tabular}
\end{table*}

\section{KV Cache Eviction on Long-Context Tasks}
\label{appendix:h2o_snap_longbench}

\begin{figure}[!ht]
    \centering
        \includegraphics[width=\linewidth]{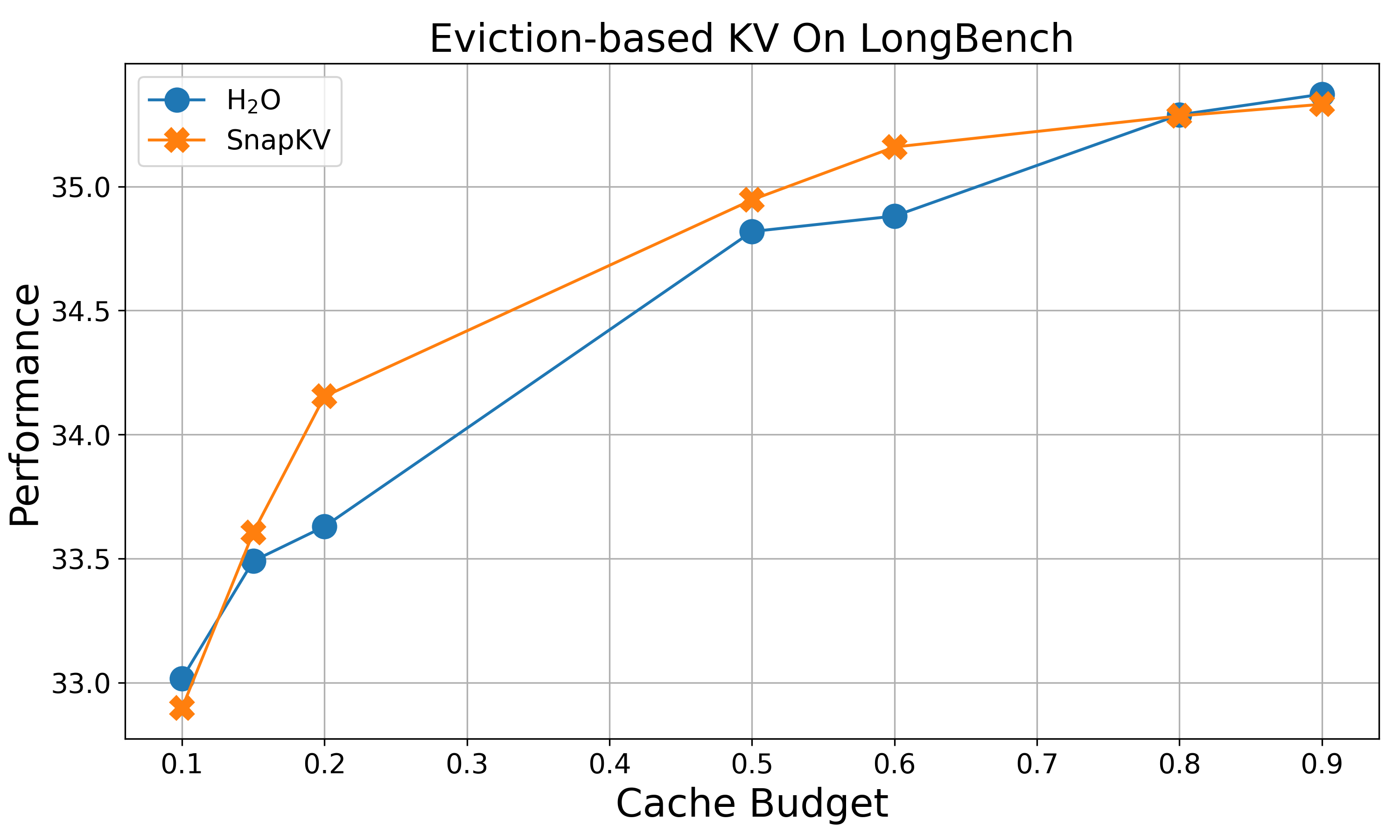}
    \caption{Eviction-based KV on LongBench: High levels of KV eviction (e.g., 80-95\%) hurts LLM's performance on long context tasks significantly. 
    }
    \label{fig:h2o_on_longbench}
\end{figure}

\fref{fig:h2o_on_longbench} shows that with 50\% KV cache size, the LLM can still obtain comparable accuracy (e.g., $<$1 point) as the full KV cache. However, high levels of KV
eviction (e.g., 80-95\%) hurts LLM’s performance on long context tasks significantly.

\section{Additional Results on Attention Distribution on Long-Context Understanding Tasks}
\label{appendix:attn}
Researchers have always been interested in exploiting the underlying structure of the attention mechanism to improve inference efficiency ~\cite{multihead-vs-singlehead,prune-head,retrieval-head}. 

\begin{figure*}[!ht]
    \centering
    \includegraphics[width=\linewidth]{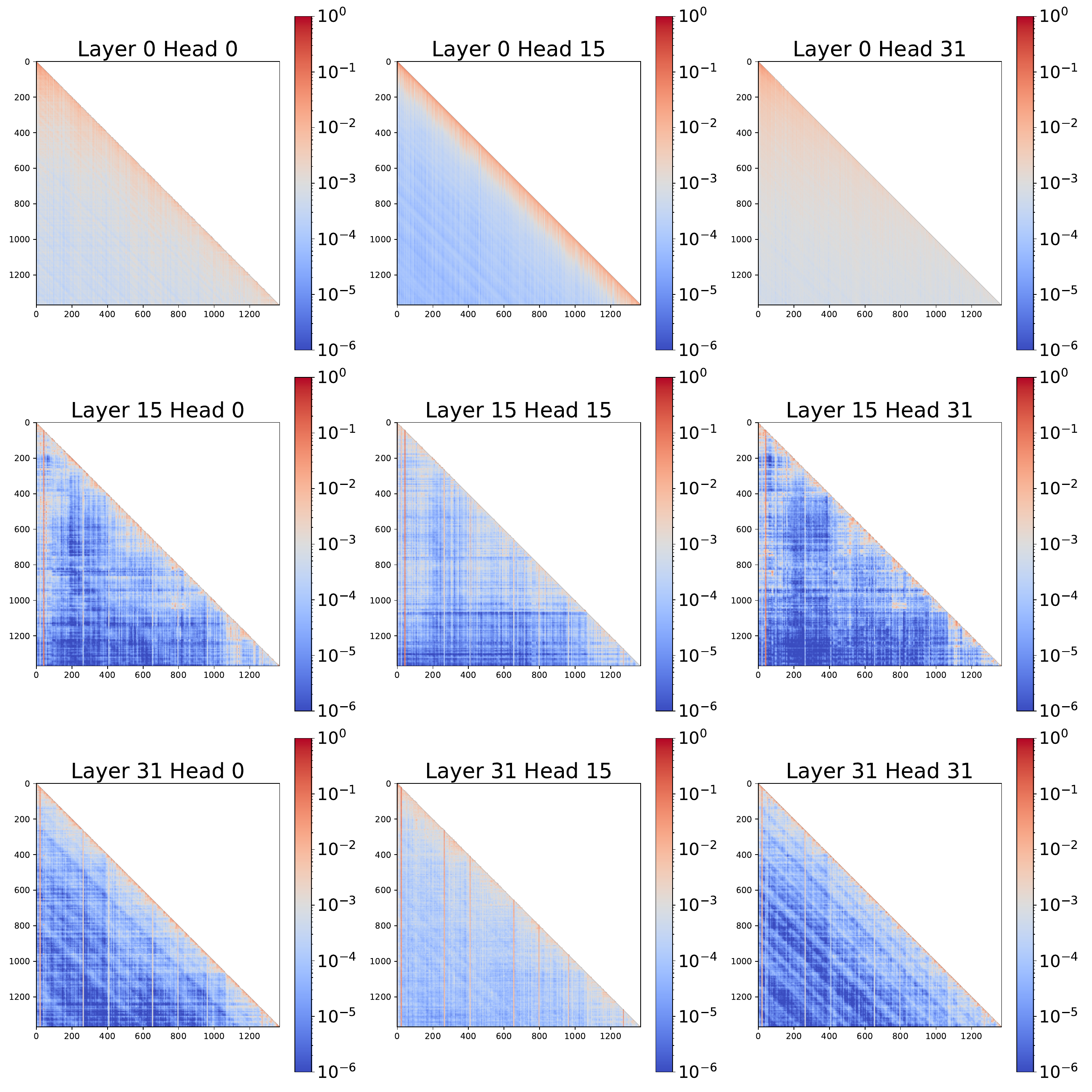}
    \label{fig:attn_plot}
    \caption{The attention distribution of LLaMA2-7B over the HotpotQA dataset in LongBench.}
    \label{fig:attn_dist_3x3}
\end{figure*}

While prior studies show that attention scores are largely sparse~\cite{h2o,attention-sink,scissorhands}, we observe that the attention distribution has more diverse patterns on long sequences. \fref{fig:attn_dist_3x3} shows that attention distribution of LLaMA2-7B-chat on a sample from the HotpotQA dataset.

We observe distinctive patterns: (i) the attention distribution at the lower layers has a wide coverage over sequence lengths and is more dispersed, and (ii) attention becomes more narrowly focused on a small subset of tokens and starts to exhibit block-wise sparse attention as the tokens move to the higher layers.  We consistently observe this pattern across datasets in LongBench. 

\section{Persistent Context Selection Analysis}
\label{appendix:persistent}
\begin{figure}[ht!]
    \centering
    \includegraphics[width=1.0\linewidth]{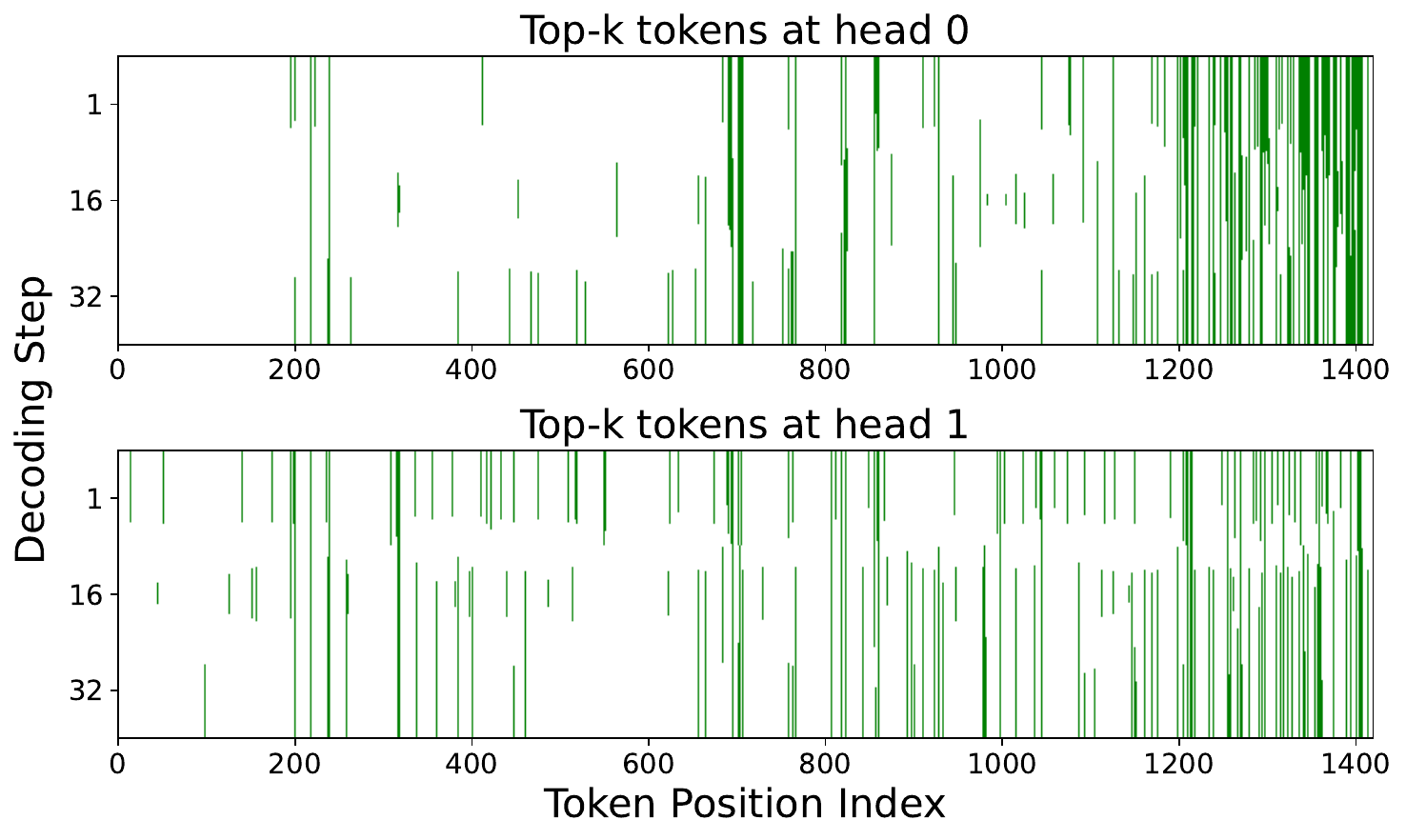}
    \caption{Top-$k$ tokens with the highest cumulative attention score on the Lcc dataset from LongBench. Green tokens mark the heavy hitters retained by the \ho algorithm. Here, we choose $k=150$.}
    \label{fig:top150}
\end{figure}

We analyzed a sample prompt from the Lcc dataset to show that the heavy hitters selected in the prefill phase persist across generations \fref{fig:top150}. The green positions indicate that the 150 heavy hitters currently retained by the \ho algorithm, while the white ones represent evicted tokens. It is evident that while different heads have different importance distributions, the important tokens largely do not vary across different generation steps.

\section{Token-Wise Quantization Of The KV Cache}
\label{appendix:token_wise_quant}
\begin{figure}[h!]
    \centering
    \includegraphics[width=1.0\linewidth]{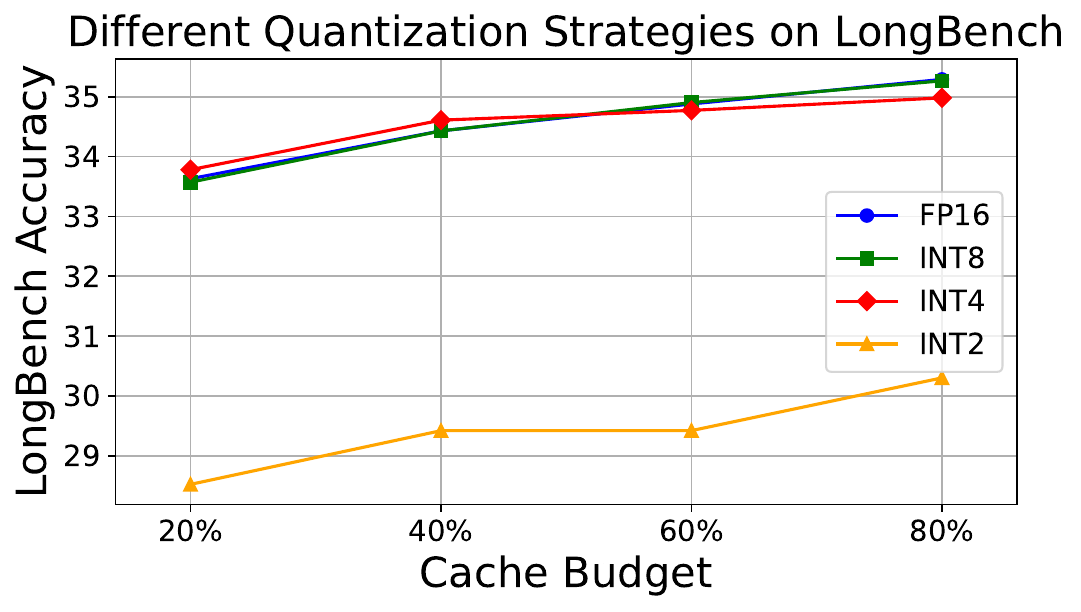}
    \caption{Performance of per-token quantized \ho on the LongBench dataset. INT8/4 quantization can maintain performance across cache budgets. However, INT2 quantization suffers from a catastrophic drop in performance.}
    \label{fig:quantised_h2o}
\end{figure}

A prevalent approach to compress the KV cache is by quantization. However, directly applying quantization to selective KV imposes challenges. 
Prior studies find that KV states contain outliers~\cite{llm-qat,smoothquant}, and per-token quantization is needed to avoid accuracy degradation. \fref{fig:quantised_h2o} shows that while applying INT8 and INT4 per-token quantization to both key and value caches helps maintain the accuracy of selective KV on LongBench, further reducing it to INT2 results in a significant accuracy drop, because 2-bits can not fully capture the dynamic range of KV token distributions. This motivates using channel-wise quantization as in KIVI \cite{kivi}
and KVQuant \cite{kvquant}.

\section{Dataset Details}
\label{appenxi:dataset}
We seek a dataset that covers a broad range of long-context understanding tasks. For this reason, we choose LongBench, which covers six major task categories and in total 13 datasets~\cite{longbench}: Qasper(F1) and MultiFieldQA(F1) are single doc QA tasks; Passage Retrieval(accuracy) and passage count(accuracy) are synthetic datasets to test the model's tendency to forgot information over a long context understanding; LCC(similarity) and RepoBench-P(similarity) are code completion tasks; 2WikiMultihopQA(F1) and HotpotQA(F1) are multi doc QA tasks; GovReport(Rouge) and MultiNews(Rouge) are summarization tasks; TREC(accuracy), SAMSum(Rouge) and TriviaQA(F1) are few-shot learning tasks. 

\section{Evaluation Details}
\label{sec::detail}

\paragraph{Decoding Strategy}
All models generate responses using deterministic greedy decoding across all tasks to ensure a fair comparison and reproducibility.


\paragraph{LongBench Truncation Strategy:}  
we ensure that the model consistently selects the first 2000 and last 2000 tokens, regardless of changes to truncation settings or special tokens. This ensures stable score calculations across tests.

\paragraph{Pyramid-like Allocation Details}
Inspired by PyramidKV\cite{pyramidkv}, we adjust the heavy hitter cache budget across layers by allocating more cache in lower layers and less in higher ones. The token allocation across layers follows a linear function. Specifically, considering the average heavy budget size is $x$, we choose a hyper-parameter pyramid depth $d$ to adjust the ratio.
The bottom-most layer has a heavy budget size of $x / d$, and the top-most layer has a heavy budget size of $2x - x / d$ with intermediate layers linearly interpolated between these values.
We choose pyramid depth $d=7$ for our experiments.

\section{KV Cache Compression Ratio Analysis}
\label{sec:kv_mem_formulas}
Given a model with $(H)$ layers, hidden dimension $(d)$, number of attention heads $(n_{heads})$, and a prompt and generated sequence of length $(l_{\text{prompt}}, l_{\text{gen}})$ the KV cache size for different techniques is shown below:

\begin{enumerate}
    \item \textbf{Full model}: All tokens are stored in FP16 format. Therefore the KV cache has size $= 2 \times (H \times d) \times (l_{\text{prompt}} + l_{\text{gen}}) \times 2\textrm{ bytes}$.
    \item \textbf{\ho}: Given a cache budget of $(\alpha_{HH}, \alpha_{RW})$ for the heavy hitters and recent window the KV cache has size $= 2 \times (H \times d) \times (l_{\text{prompt}}) \times (\alpha_{HH} + \alpha_{RW}) \times 2\textrm{ bytes}$
    \item \textbf{\snap}: Given a cache budget of $p$, \snap performs eviction in the prefill phase and retains all generated tokens. Hence, the KV cache has size $= 2 \times (H \times d) \times (p*l_p + l_g) \times 2\textrm{ bytes} $
    
    \item \textbf{KIVI}: With a group size of 16, i.e., 16 scalars quantized from FP16 to INT2 format, the memory required by a group is 16 scalars $\times 2$ bits = 4 bytes. The quantization zero-point and scale are saved in FP16 format and require $2 \times 2$ bytes. In total, the group requires 8 bytes. Hence, the KV cache has $(H \times d) \times (l_{\text{prompt}} + l_{\text{gen}}) \textrm{ bytes}$.
    \item \textbf{Q-Hitter}: The Q-hitter paper performs INT4 token quantization per attention head. Therefore, the $(d/n_{heads})$ scalars which would be stored in FP16 are now stored in 4-bit precision. The quantization metadata is the zero-point and scale, both in FP16 precision. Therefore, the compression factor for Q-Hitter is $(d/n_{heads} * 16) / (d/n_{heads} * 4 + 2*16)$. For the Llama-7B-chat model this number is $(4096/32 * 16)/(4096/32*4 + 32) = 3.76 \times$. Hence, the KV cache size is $2 \times (H \times d) \times (l_{\text{prompt}}) \times (\alpha_{HH} + \alpha_{RW}) \times 2 / 3.76 \textrm{ bytes}$
    
    \item \textbf{\name}: The prompt tokens are evicted with a cache budget of $\alpha_{HH}, \alpha_{RW}$ and all generated tokens are retained. All tokens are stored in 2-bit precision.
    Similar to KIVI, each group of 16 scalars and their quantization metadata requires $8$ bytes in total.
    Hence, the size of the KV cache is $ = (H \times d) \times (\alpha_{HH} + \alpha_{RW}) \times (l_{\text{prompt}}) + (H \times d) \times (l_{\text{gen}})\textrm{ bytes}$.
 \end{enumerate}

Given a certain prompt and output length, the uncompressed baseline and KIVI have a fixed KV cache size. However, \ho, Q-Hitter, and \name can tune the cache budget $(\alpha_{HH}, \alpha_{RW})$ to modify the KV cache size.

For prompt length $4096$ and generation length $512$ the full model's and \name 's KV cache consume 2.4GB and 0.33GB respectively. Therefore, \name leads to an $(1-0.33/2.4) = 86\%$ reduction in KV cache size.

\section{Performance against KV cache size}
\label{appendix:perf_vs_kv_size}
\begin{figure*}[!ht]
    \centering
    \begin{minipage}[b]{0.45\textwidth}
        \centering
        \includegraphics[width=\textwidth]{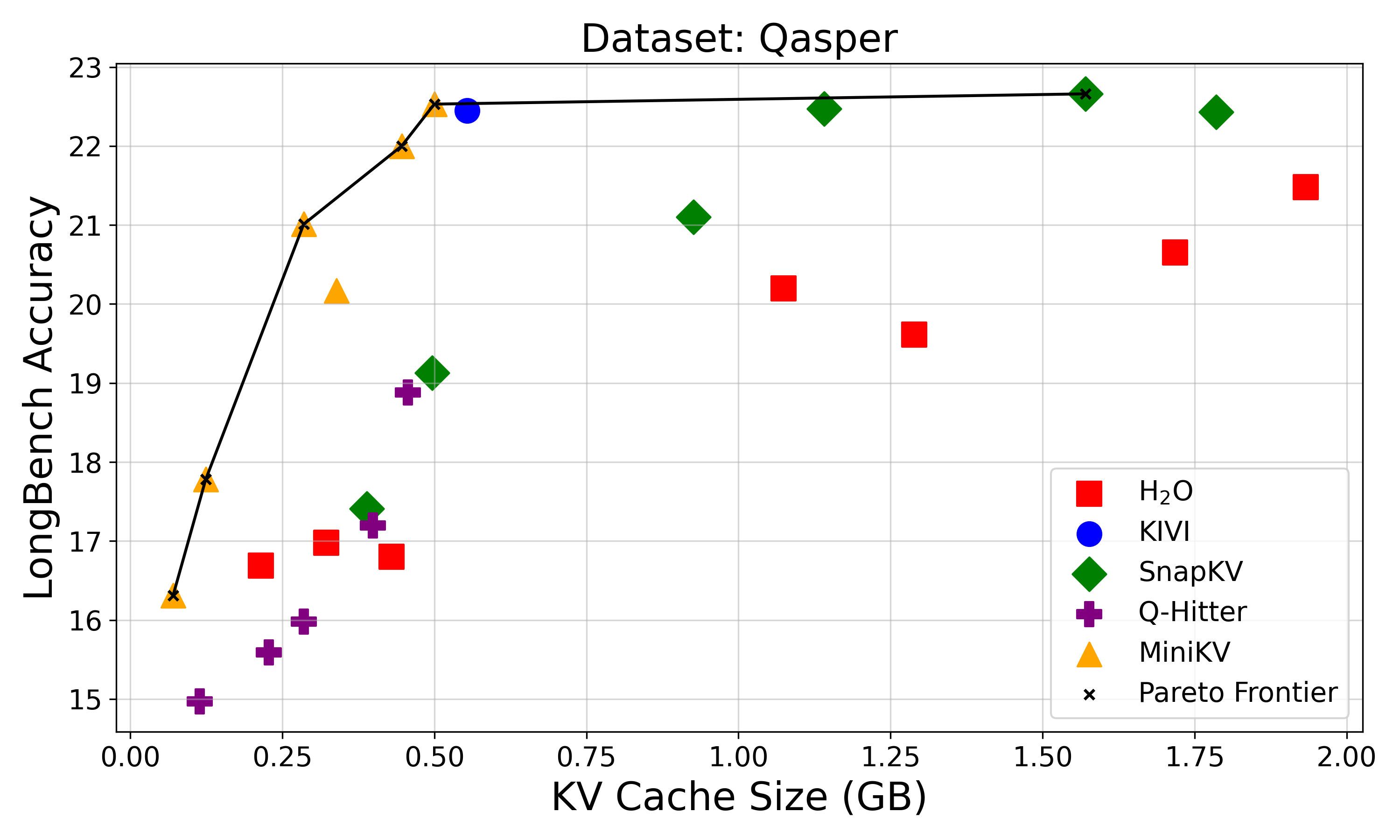}
    \end{minipage}
    \begin{minipage}[b]{0.45\textwidth}
        \centering
        \includegraphics[width=\textwidth]{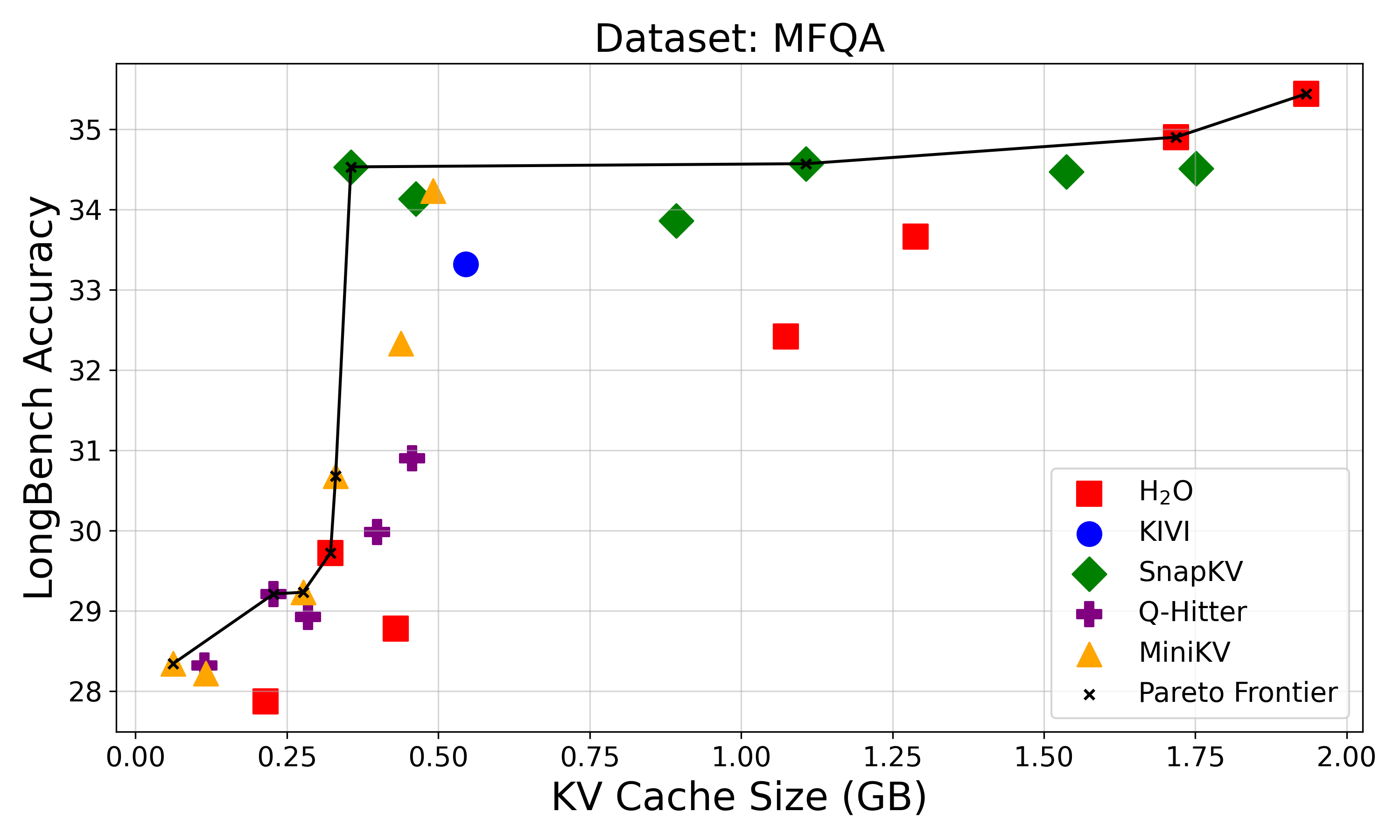}
    \end{minipage}
    \begin{minipage}[b]{0.45\textwidth}
        \centering
        \includegraphics[width=\textwidth]{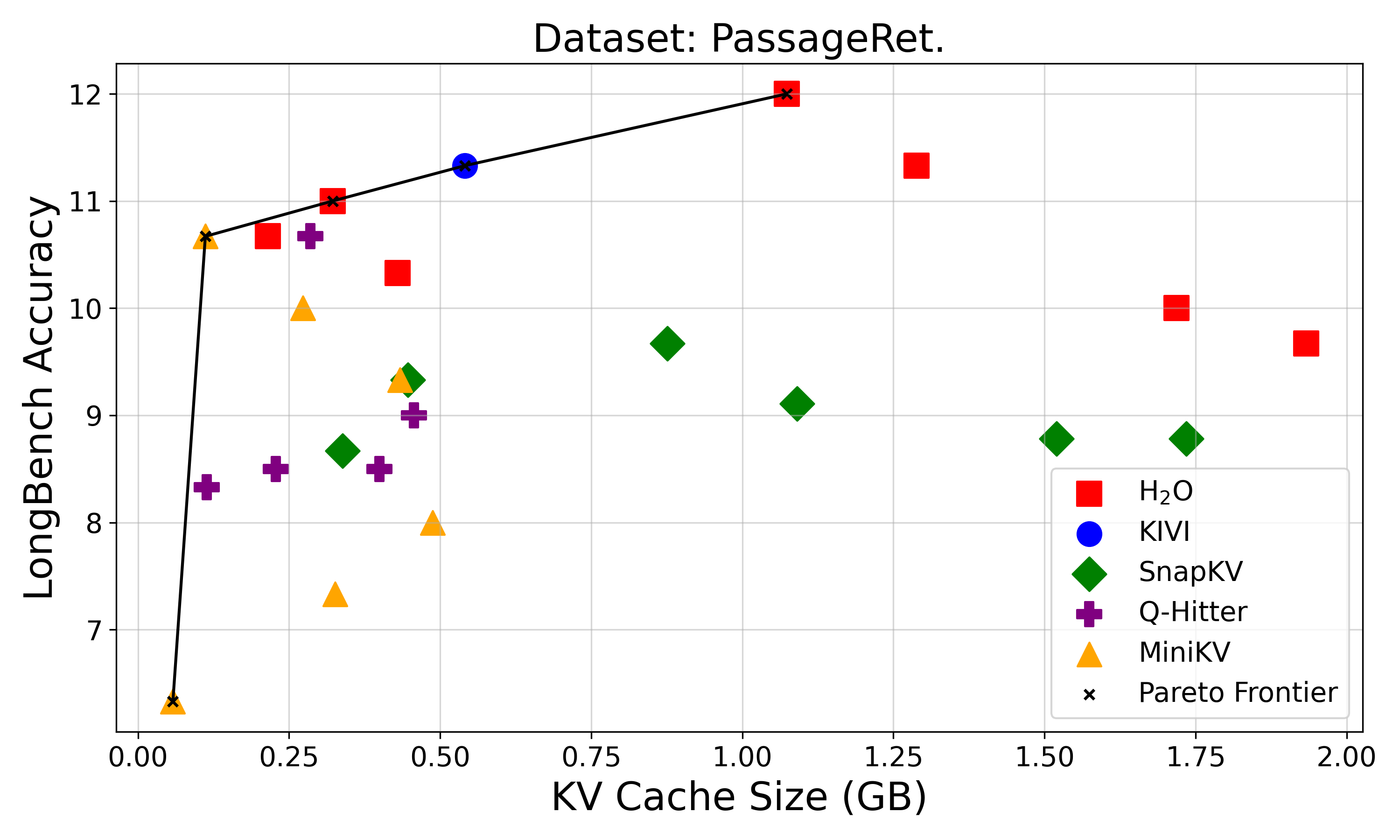}
    \end{minipage}
    \begin{minipage}[b]{0.45\textwidth}
        \centering
        \includegraphics[width=\textwidth]{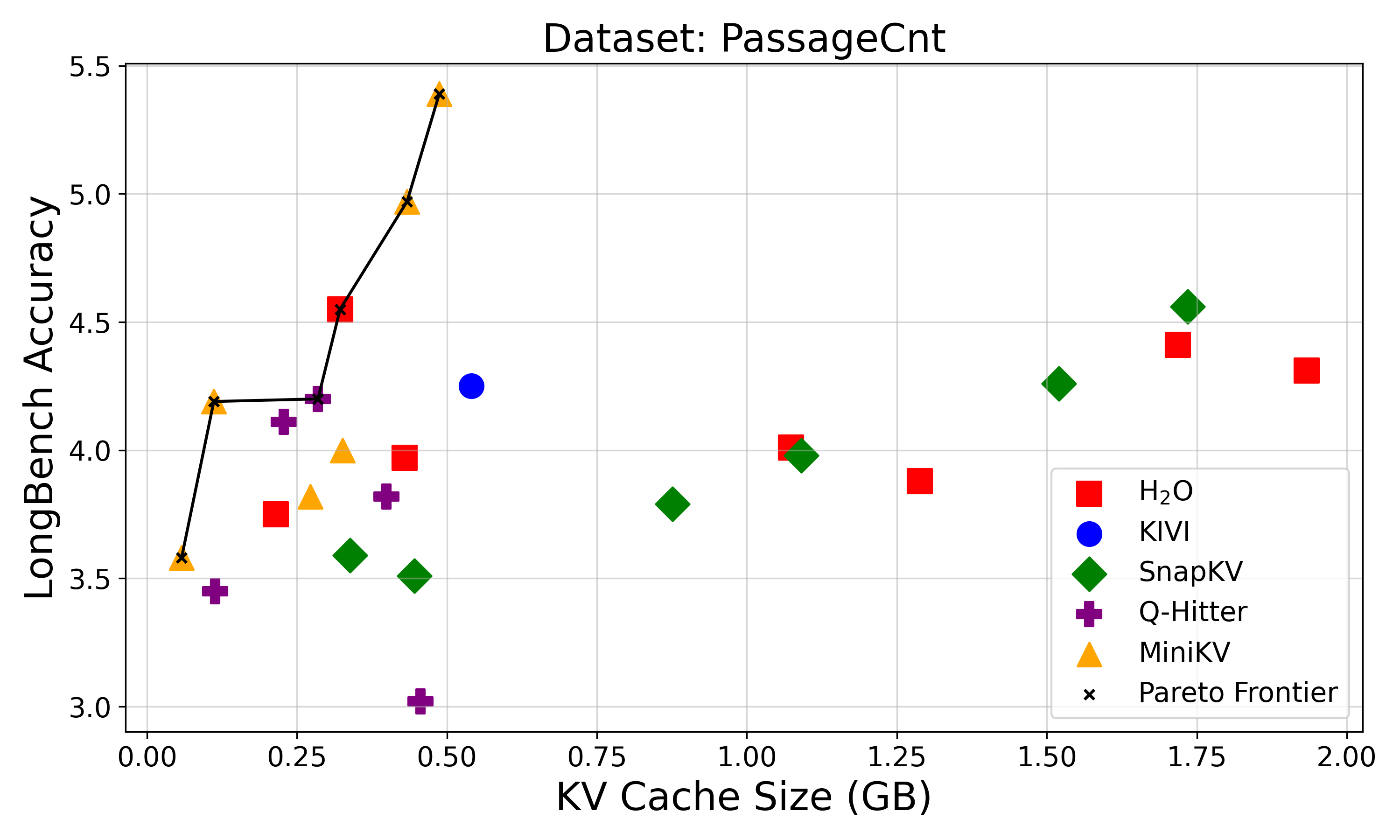}
    \end{minipage}
    \begin{minipage}[b]{0.45\textwidth}
        \centering
        \includegraphics[width=\textwidth]{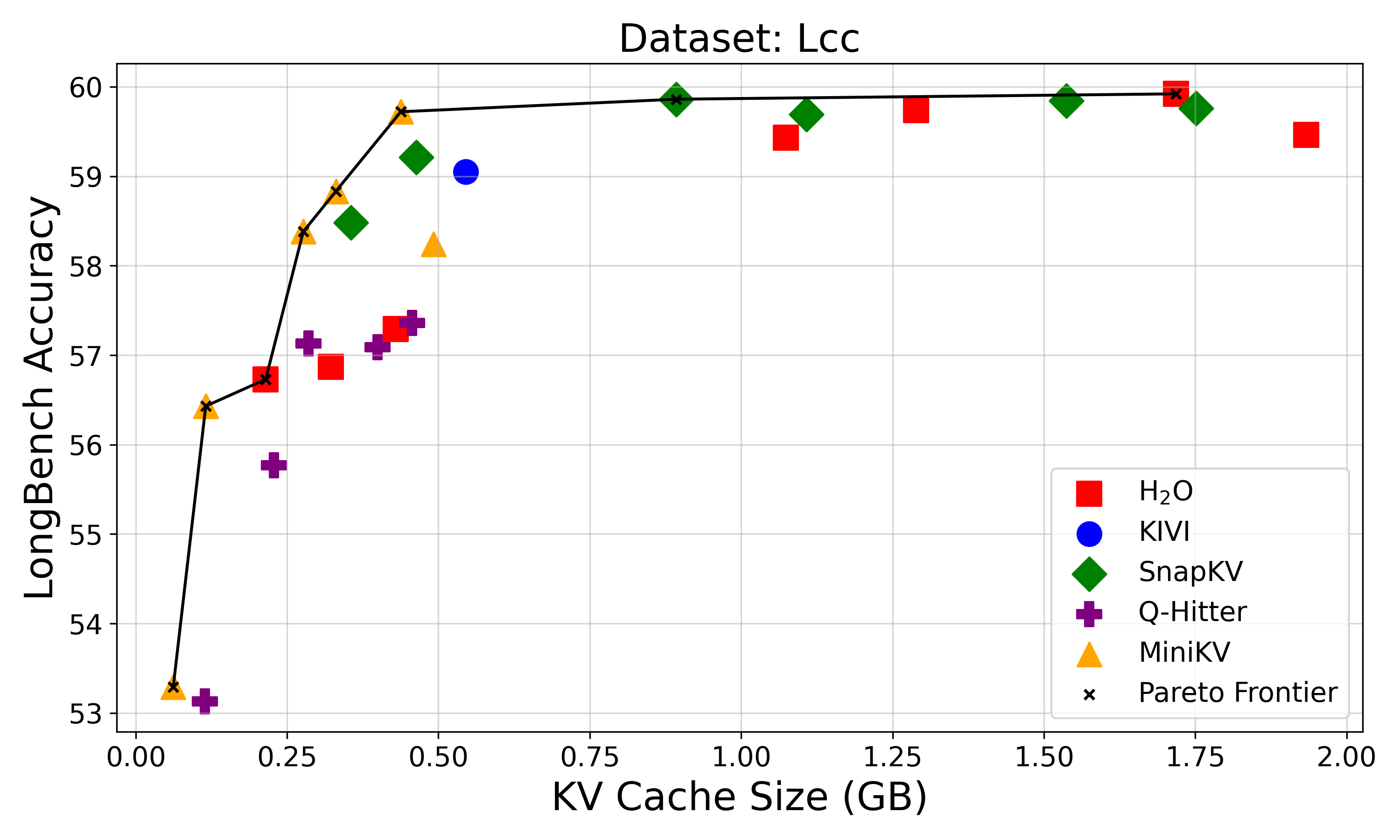}
    \end{minipage}
    \begin{minipage}[b]{0.45\textwidth}
        \centering
        \includegraphics[width=\textwidth]{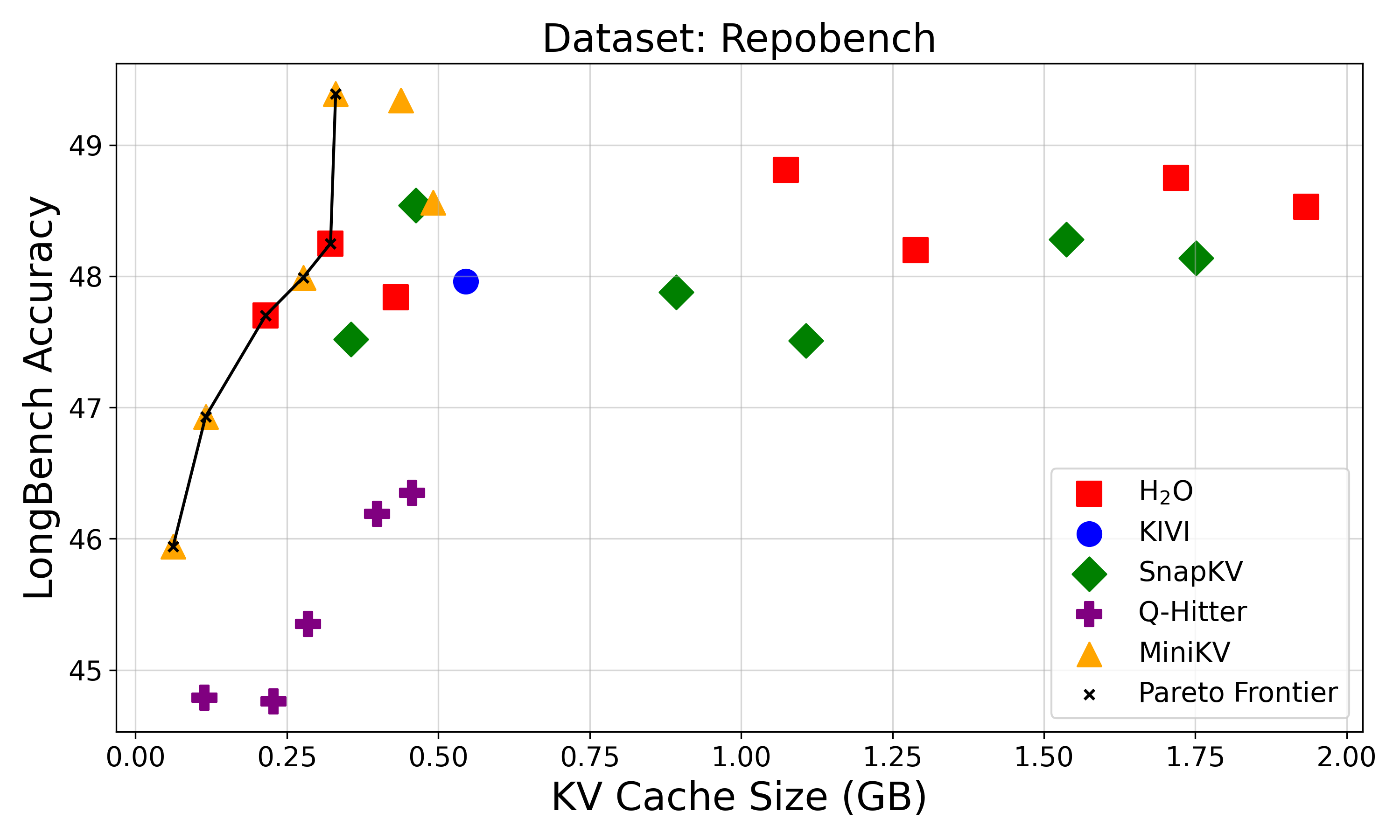}
    \end{minipage}
    \begin{minipage}[b]{0.45\textwidth}
        \centering
        \includegraphics[width=\textwidth]{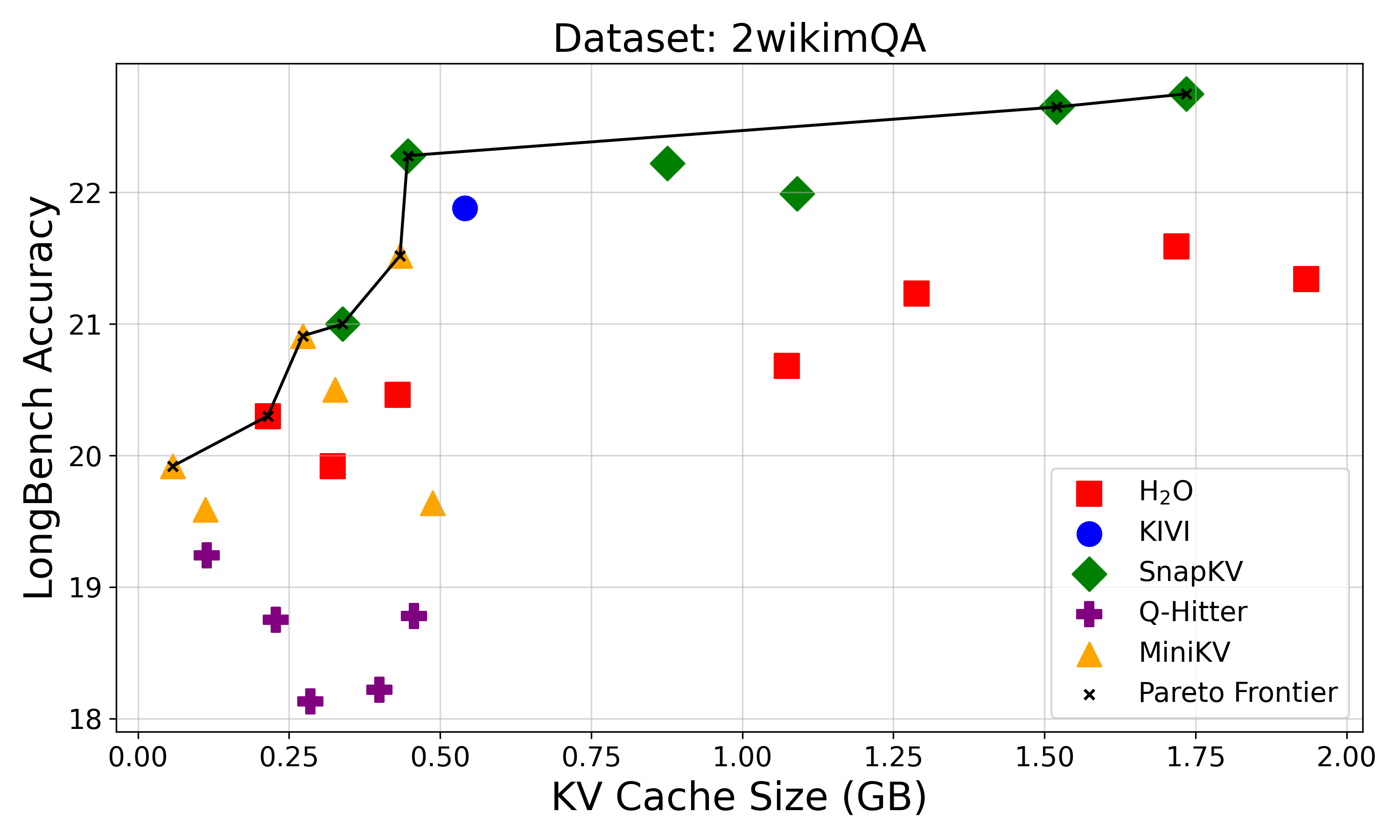}
    \end{minipage}
    \begin{minipage}[b]{0.45\textwidth}
        \centering
        \includegraphics[width=\textwidth]{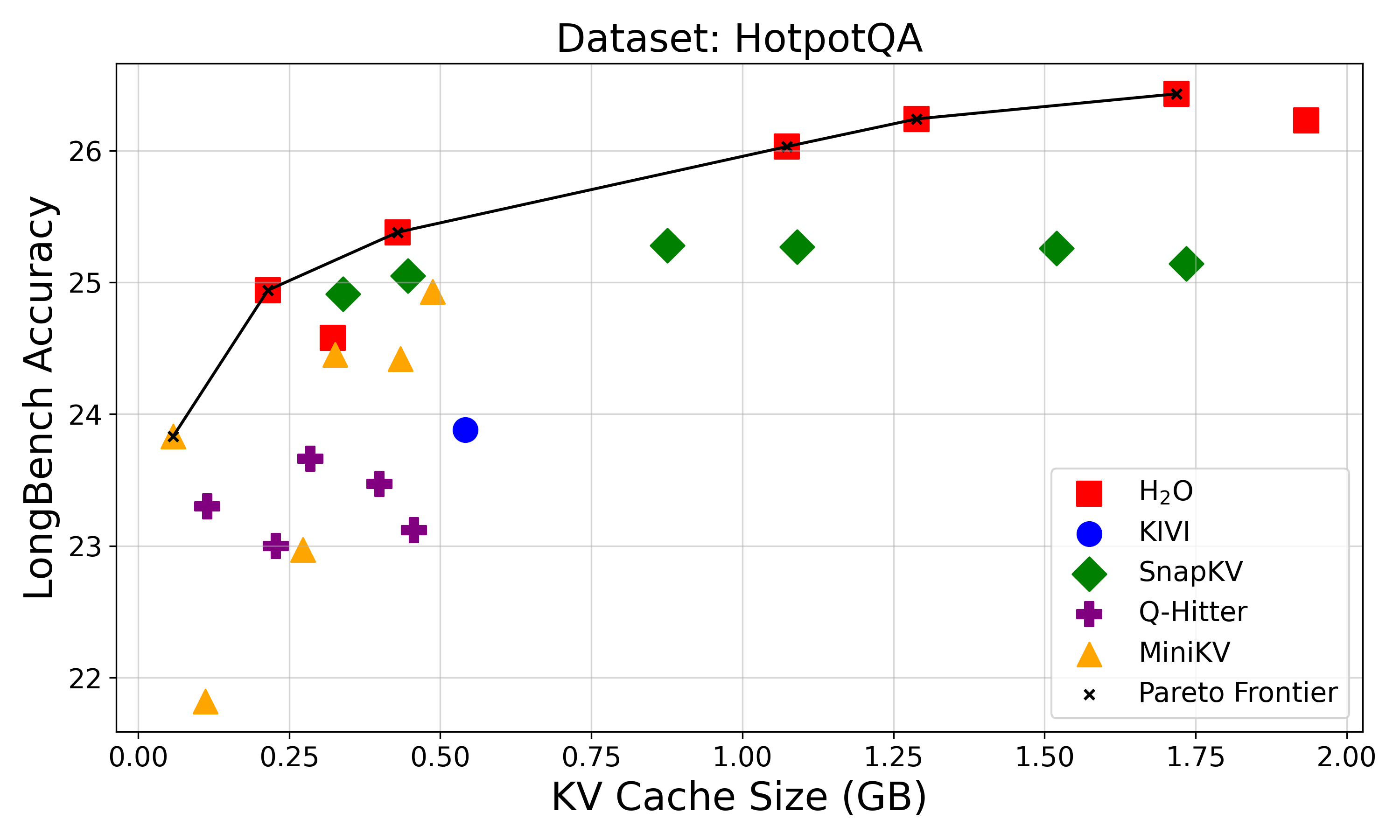}
    \end{minipage}
    
    \caption{Performance Versus KV Cache Size: \name offers the best performance for the smallest KV cache size across all 6 task categories.}
    \label{fig:appendix_perf_vs_kv_size_1}
\end{figure*}

\begin{figure*}[!ht]
    \centering
    \begin{minipage}[b]{0.45\textwidth}
        \centering
        \includegraphics[width=\textwidth]{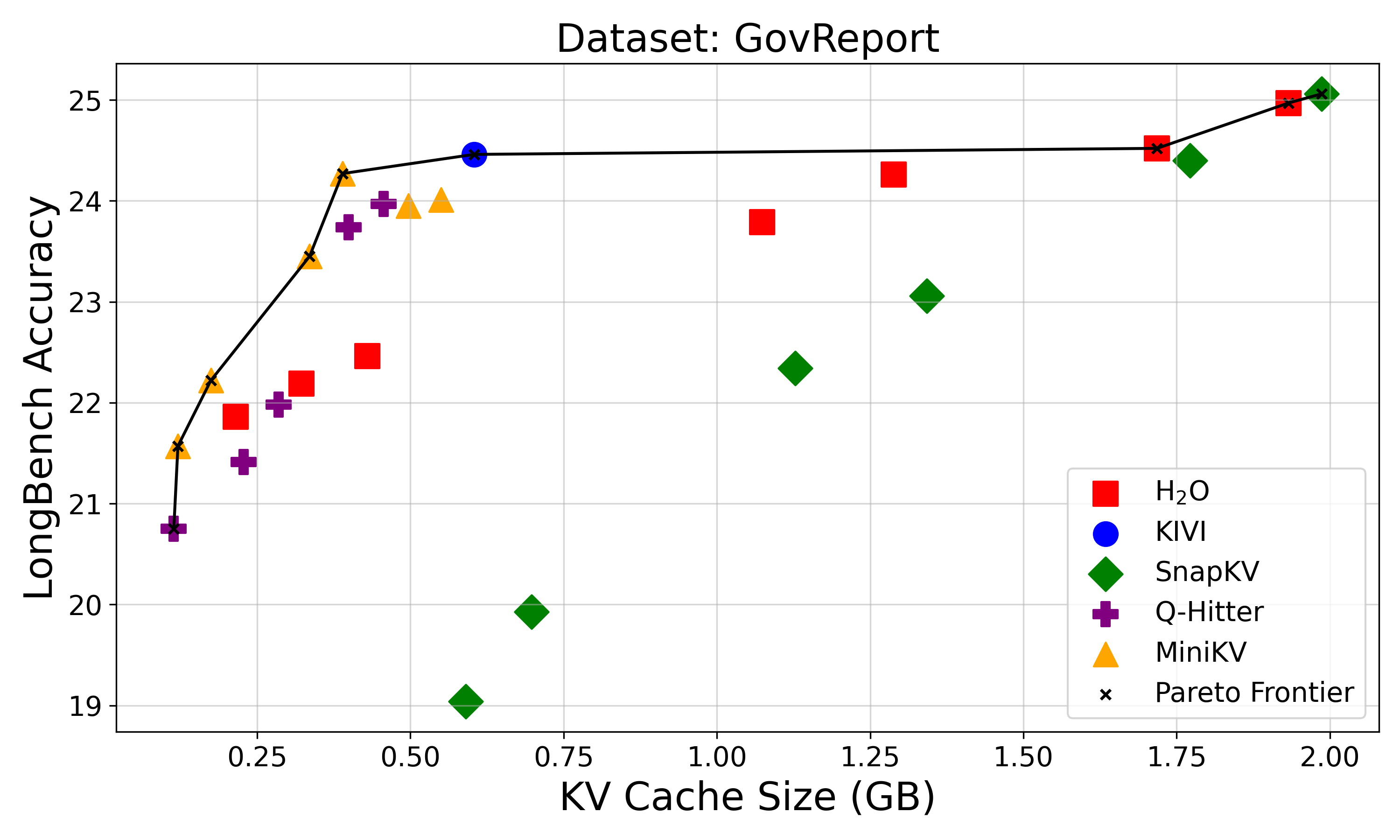}
    \end{minipage}
    \begin{minipage}[b]{0.45\textwidth}
        \centering
        \includegraphics[width=\textwidth]{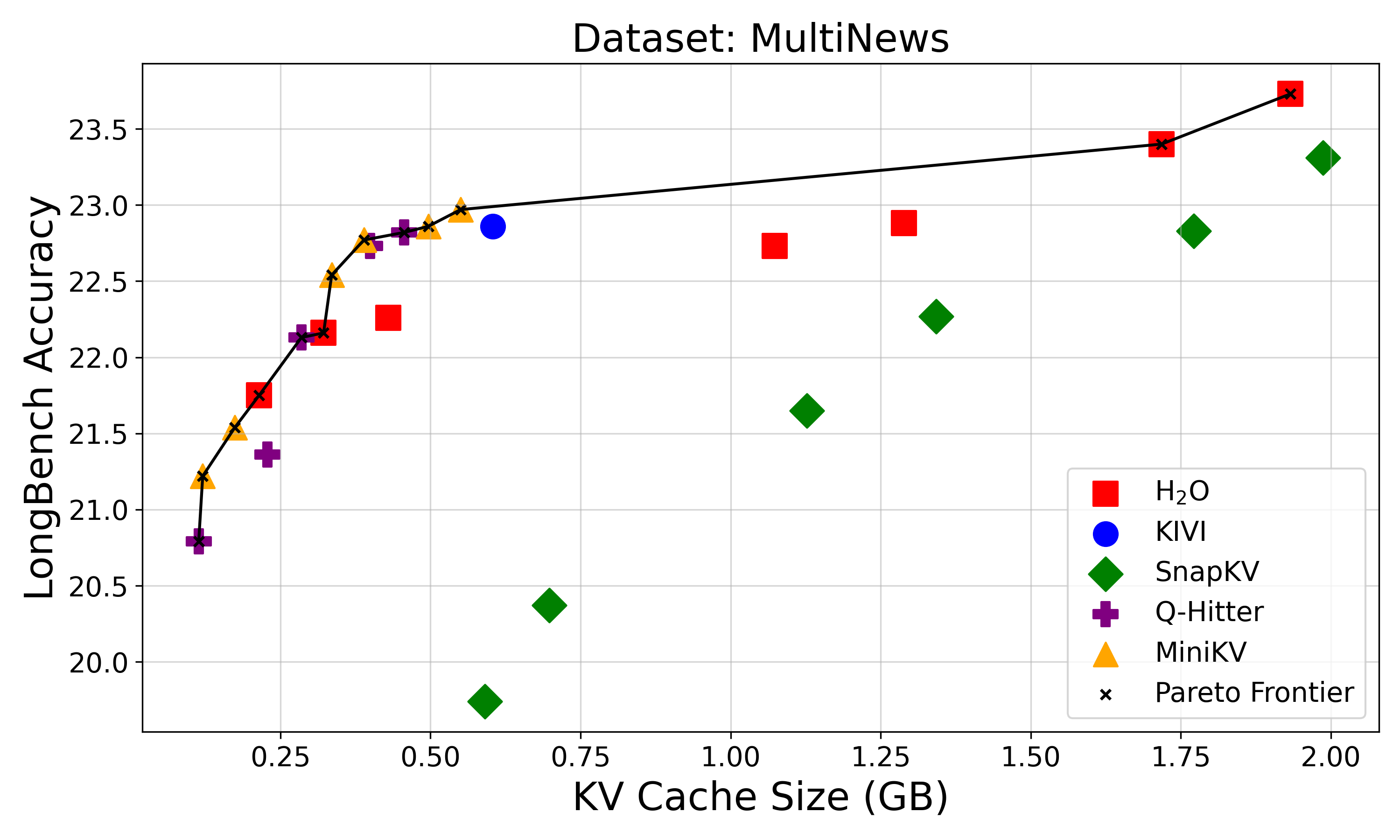}
    \end{minipage}
    \begin{minipage}[b]{0.45\textwidth}
        \centering
        \includegraphics[width=\textwidth]{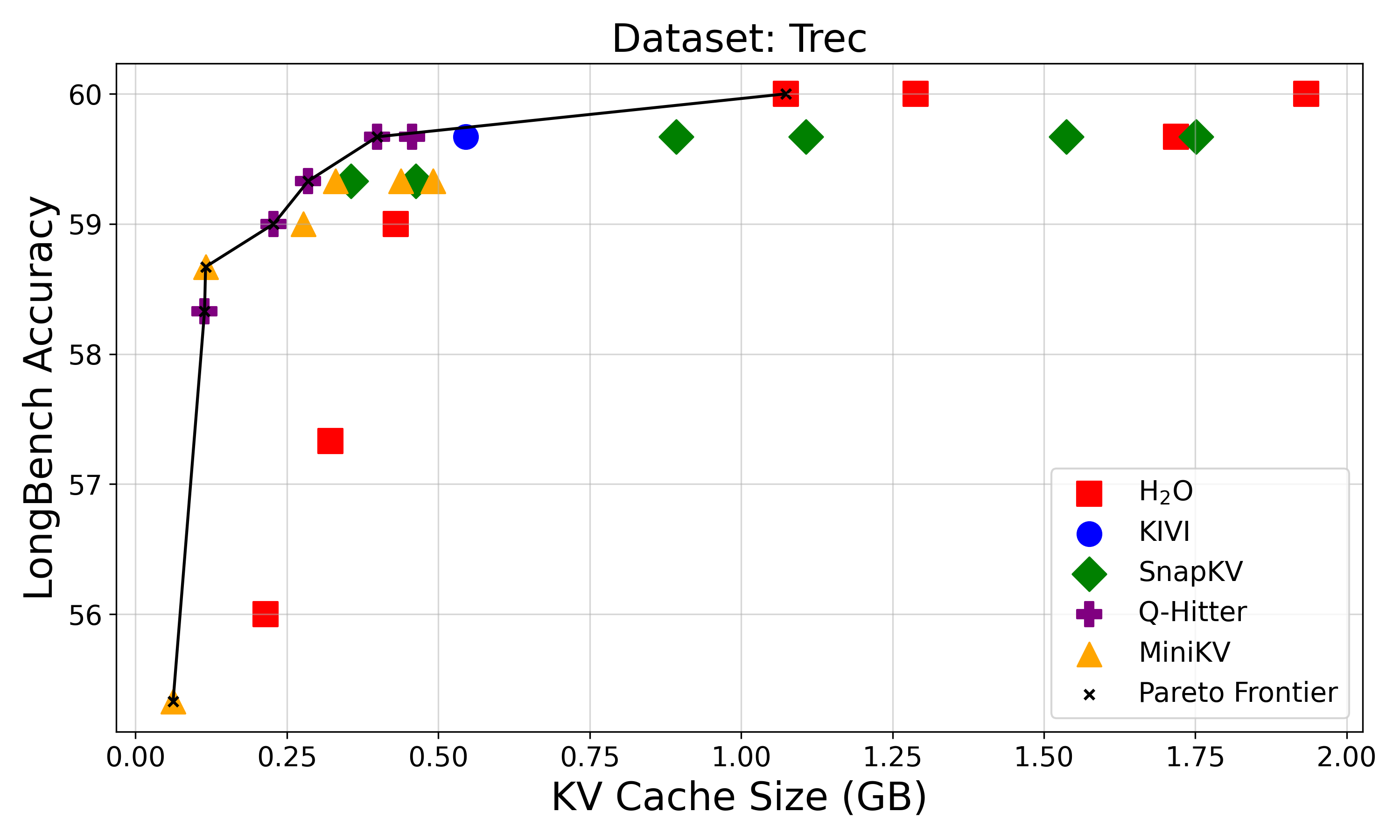}
    \end{minipage}
    \begin{minipage}[b]{0.45\textwidth}
        \centering
        \includegraphics[width=\textwidth]{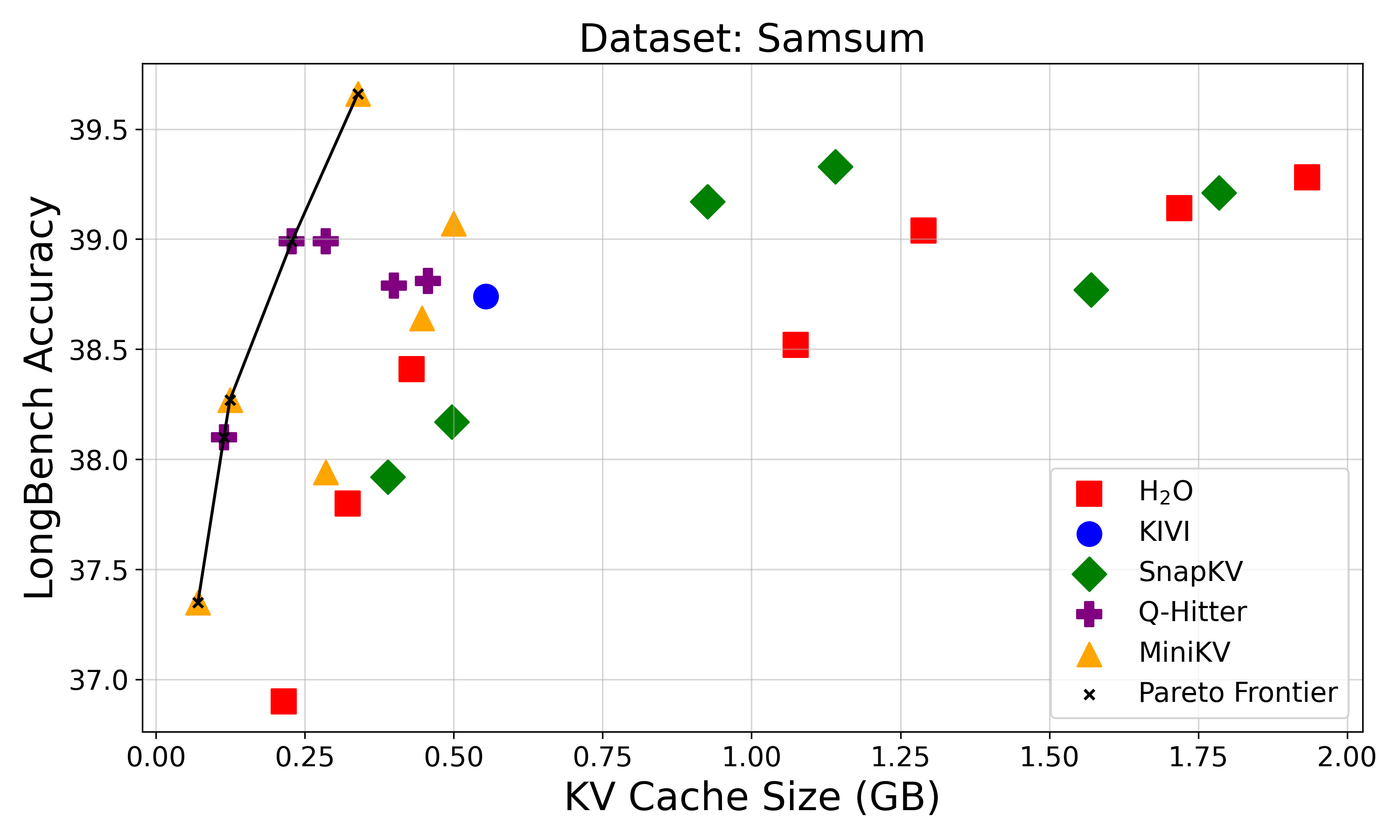}
    \end{minipage}
    \begin{minipage}[b]{0.45\textwidth}
        \centering
        \includegraphics[width=\textwidth]{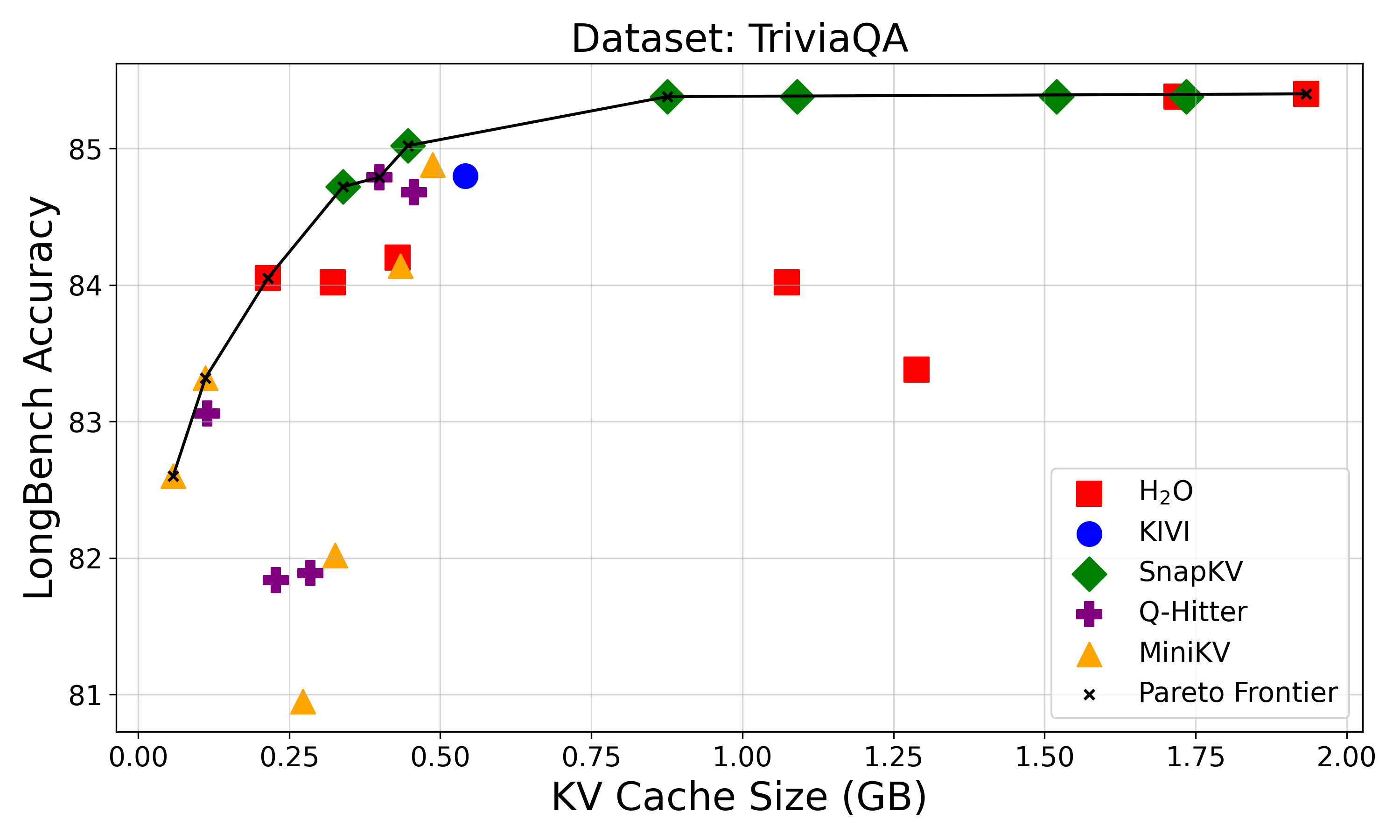}
    \end{minipage}
    \caption{Performance Versus KV Cache Size: \name offers the best performance for the smallest KV cache size across all 6 task categories.}
    \label{fig:appendix_perf_vs_kv_size_2}
\end{figure*}

As discussed in \sref{sec:kv_mem_formulas}, the KV cache size depends on the prompt and generation length. Each dataset in LongBench has a different maximum generation length, therefore we make separate plots for each dataset with prompt length $4096$ and the generation length as the dataset-specific maximum generation length.
Figure \ref{fig:appendix_perf_vs_kv_size_1} and \ref{fig:appendix_perf_vs_kv_size_2} show the performance vs KV cache size curve. \name achieves the optimal compression strategy across all six major task categories on LongBench (single/multi-doc QA, LC understanding, code completion, summarization, and few-shot learning). These results validate the effectiveness of \name with varying KV cache sizes.

\section{End-To-End Latency Breakdown}
\label{appendix:e2e-breakdown}
\begin{figure}[H]
    \centering
    \includegraphics[width=0.6\columnwidth]{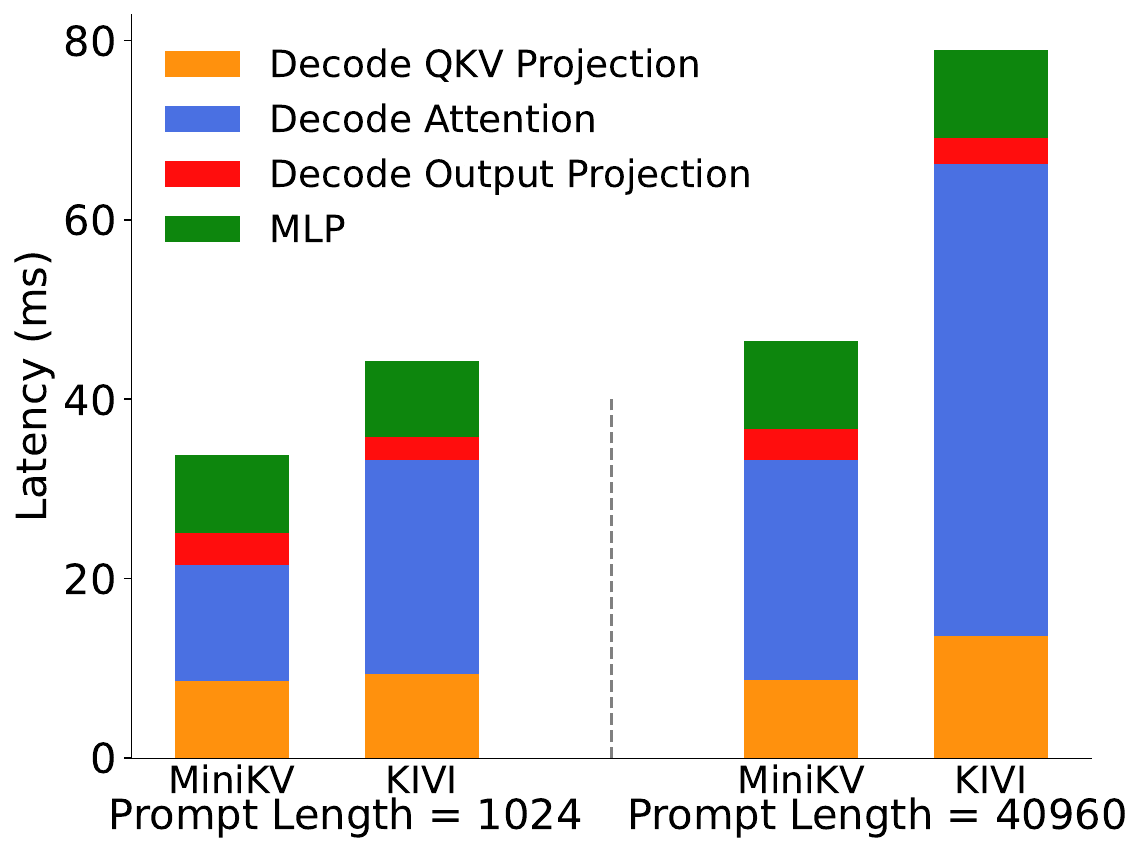}
    \caption{Per token latency breakdown for the decoding phase. Generation length = 1024 and batch size = 1.}
    \label{fig:latency_breakdown}
\end{figure}

We analyze the breakdown of latency associated with each computation in the standard decoder layer of the transformer architecture for \name and KIVI during the decoding phase. We particularly look at latencies for projections of the input vector into query, key, and value vectors, attention computation, and output projection. We also measure the time spent in the MLP layer. We present the latency breakdown as the total latency for each computation component divided by the generation length.

As shown in \fref{fig:latency_breakdown}, \name achieves a lower end-to-end latency than KIVI. This improvement primarily arises during attention computation as well as projection of Query, Key and Value. Specifically, the inference time is dominated by KV cache loading time when processing long contexts. Therefore, \name's smaller KV cache results in reduced KV load times from the GPU’s HBM.

\end{document}